\documentclass{article} 
\usepackage{iclr2026_conference,times}
\usepackage{graphicx}

\usepackage{amsmath,amsfonts,bm}

\def\eqref#1{equation~\ref{#1}}

\def\1{\bm{1}}

\DeclareMathAlphabet{\mathsfit}{\encodingdefault}{\sfdefault}{m}{sl}
\SetMathAlphabet{\mathsfit}{bold}{\encodingdefault}{\sfdefault}{bx}{n}

\usepackage{hyperref}
\usepackage{url}
\def\eg{{\textit{e.g.}}}

\newcommand{\ryn}[1]{\textcolor{black}{#1}}

\newcommand{\qzf}[1]{\textcolor{black}{#1}}

\title{UMI3D: Robust 3D Generation on Unconstrained Multi-Image Inputs via Simultaneous Focus Cross-Attention Routing}

\author{Zefan Qu, Zhenwei Wang, Gerhard Petrus Hancke, Rynson W.H. Lau \\
Department of Computer Science\\
City University of Hong Kong\\
\texttt{\{zefanqu2-c, zhenwwang2-c\}@my.cityu.edu.hk} \\
\texttt{\{gp.hancke, rynson.lau\}@cityu.edu.hk} \\
}

\iclrfinalcopy 
\begin{document}

\maketitle

\begin{abstract} Recent 3D foundation models can generate high-quality assets from a single image, but degrade markedly on unconstrained multi-image inputs, often producing distorted geometry, over-smoothed textures, and chaotic colors. We argue that this failure stems not from limited model capacity, but from a mismatch between single-image cross-attention and the multi-image setting: existing models lack a principled way to decide which image each 3D voxel should trust at each denoising step. Revisiting recent single-image 3D foundation models, we show that explicitly routing each voxel to its most informative image is sufficient to unlock strong performance on inconsistent multi-image inputs. Based on this observation, we propose UMI3D, a training-free and plug-and-play framework that restructures cross-attention for unconstrained multi-image 3D generation. Its core, \textit{Simultaneous Focus Cross-Attention} (SFC-Attn), activates all conditioning images at each denoising step while allowing each voxel to focus on the single image that best explains it. To enable this routing, we derive the Voxel Reference Score (VRS), a model-intrinsic metric for voxel--image affinity that requires no external matching, segmentation, or correspondence models. Extensive experiments show that UMI3D unlocks the multi-image potential of single-image 3D generation frameworks across diverse tasks. Project Page: \href{UMI3D-Project.github.io}{\textcolor{blue}{UMI3D-Project.github.io}}.
\end{abstract}

\section{Introduction}

\label{intro}
Generating 3D objects from multiple input images is \ryn{a common approach}
for many real-world 3D workflows, such as creating 3D characters from multi-view concept art and constructing virtual environments from collections of scene photographs. In practice, although \ryn{the multiple input images} depict the same underlying object, they often differ in pose, occlusion, style, local details, and illumination, frequently violating the strict 3D consistency assumptions required by traditional \ryn{multi-view} to 3D methods. We refer to such image collections as \emph{unconstrained multi-image
inputs}, and the corresponding task of producing a coherent 3D asset
from them as \emph{unconstrained multi-image 3D generation}.

\begin{figure}
\vspace{-5mm}
  \includegraphics[width=\textwidth]{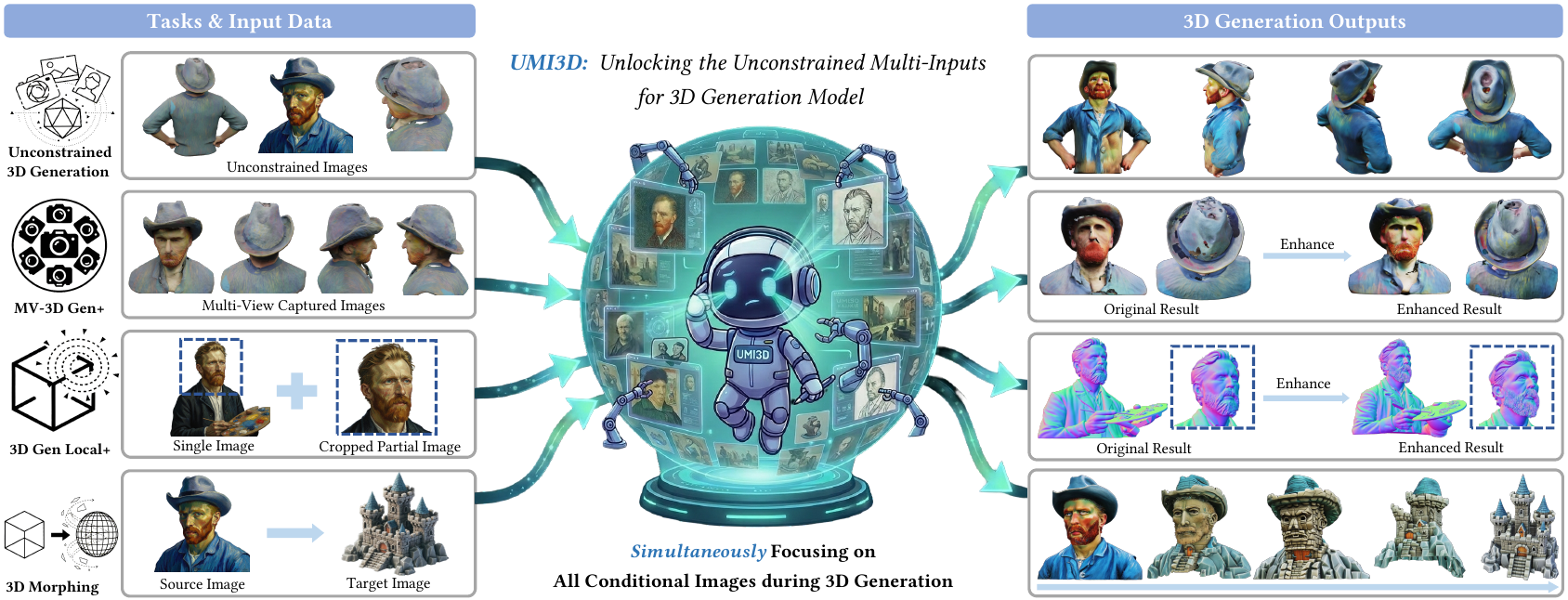}
  \caption{
UMI3D can generate a coherent 3D asset with unconstrained multi-image input
and support multiple downstream applications in a training-free manner.
}
  \label{fig:teaser}
\end{figure}

Recently, 3D \ryn{generative} foundation models~\citep{li2025sparc3d,wu2025direct3d, hunyuan3d2025hunyuan3d,xiang2025native,xiang2024structured, li2025triposg,chen2025sam} have made remarkable progress, enabling high-quality 3D asset generation from a single image. However, most existing 3D generation frameworks are designed primarily for single-image input. When naively extended to multi-image inputs, these models often exhibit severe degradation, including distorted geometry, over-smoothed or inconsistent textures, and color inconsistencies across views. This gap motivates us to revisit how single-image-oriented 3D foundation models integrate information from multiple inputs.


Some existing methods~\citep{hunyuan3d2025hunyuan3d,li2025triposg} incorporate multi-image conditioning during training, but are designed for relatively consistent multi-view observations. When the inputs exhibit substantial discrepancies in visibility, appearance, or local details, these methods produce entangled 3D structures that limit their real-world applicability. Meanwhile, single-image 3D generation frameworks~\citep{xiang2024structured,xiang2025native,wu2025direct3d} can be naively extended to multiple inputs, for example by alternating the conditioning image across denoising steps or by combining independently generated outputs. However, such strategies force the model to switch its conditioning reference across denoising steps, often leading to detail loss, conditioning oscillation and noticeable discrepancies across views, \ryn{as shown in Fig.~\ref{fig:intro}}.

Our analysis shows that this gap traces back to the mismatch between single-image cross-attention design and the multi-image setting. Existing single-image 3D generation frameworks are built around cross-attention to a single conditioning image, but multi-image generation requires different parts of the 3D asset to extract information from different references. Without an explicit mechanism to regulate this process, visual cues from multiple images can interfere with one another, leading to unstable geometry and inconsistent textures. Yet a closer look at the cross-attention itself reveals a useful signal hidden in plain sight. The attention patterns of a voxel over an image already encode if that image truly explains the voxel, and this cue can be read out without any external supervision. This observation motivates us to revisit multi-image 3D generation from the perspective of \emph{selective visual routing}. If the model can determine which reference image is most informative for each 3D region while still leveraging all inputs during generation, then the strong 2D-3D priors in existing single-image foundation models can be effectively extended to unconstrained multi-image settings.


In this work, we propose \textbf{UMI3D}, a training-free framework for unconstrained multi-image 3D generation. At its core, UMI3D introduces a \textbf{S}imultaneous \textbf{F}ocus \textbf{C}ross-\textbf{A}ttention (SFC-Attn) module, which activates all conditioning images simultaneously at each denoising step while letting each 3D voxel focus on the image that best explains it, thereby avoiding the conditioning oscillation caused by switching references across steps. To drive this routing, we analyze the original cross-attention layers and derive a metric called \textbf{V}oxel \textbf{R}eference \textbf{S}core (VRS), which automatically establishes voxel–image correspondences without relying on any external models (e.g., 3D matching or segmentation networks). VRS exploits how the baseline 3D generation model utilizes both global and local image information, allowing reliable voxel-to-condition assignments in a purely model-intrinsic manner. By recasting cross-attention as a dynamic routing problem, UMI3D ensures that each 3D region is conditioned on the most informative input image.

Beyond unconstrained multi-image 3D generation, the same mechanism extends 
naturally to several additional settings, including 3D morphing, multi/single-view 3D generation enhancement, which we showcase as applications of our framework. To systematically evaluate the core task, we construct a benchmark of unconstrained multi-image inputs covering diverse cross-view inconsistencies, on which extensive experiments 
demonstrate that UMI3D delivers strong and robust multi-image 3D generation performance.
Our main contributions of this work can be summarized as:

\begin{itemize}
\item[$\bullet$] We propose UMI3D, a training-free framework that adapts single-image 3D foundation models to unconstrained multi-image inputs without modifying or retraining the backbone.
\item[$\bullet$] We introduce the Voxel Reference Score (VRS), a model-intrinsic metric that estimates voxel--image affinities from the behavior of the original cross-attention layers, without relying on external correspondence, matching, or segmentation models.
\item[$\bullet$] We propose the Simultaneous Focus Cross-Attention (SFC-Attn) module, which performs voxel-wise image routing within each denoising step, enabling more consistent 3D generation from multiple input images.
\item[$\bullet$] We \ryn{have constructed} a new benchmark for unconstrained multi-image 3D generation with realistic cross-view inconsistencies. Extensive experiments \ryn{on this benchmark} validate the effectiveness and robustness of UMI3D.
\end{itemize}

\section{Related Works}

\textbf{3D Generative Foundation Models}
Recently, a growing body of \ryn{works}~\citep{luo2021diffusion, nichol2022point, hui2022neural,muller2023diffrf, chen2023single, shue20233d, poole2022dreamfusion,wang2024themestation} has explored \ryn{the transfer} of 2D diffusion models~\citep{ho2020denoising, rombach2022high} to various 3D generation tasks.
To further improve generation quality and computational efficiency, another line of research~\citep{ren2024xcube, zhang20233dshape2vecset, zhang2024clay, chen2019net, hertz2022spaghetti} investigates \ryn{how to perform} 3D generation in compact latent spaces.
While several of these approaches focus predominantly on geometric modeling, they typically rely on an additional texturing stage to obtain fully usable 3D assets.
\ryn{Most} recently, the emergence of large-scale 3D generative models~\citep{zhao2025hunyuan3d,xiang2024structured,wu2024direct3d,hui2024make, li2025triposg, seed2025seed3d,chen2025sam} has further accelerated progress in 3D generation, providing stronger and more versatile priors for learning 3D content. Voxel-based 3D generative models, exemplified by TRELLIS~\citep{xiang2025native} and Direct3D~\citep{wu2025direct3d}, have recently emerged as a main paradigm for 3D generative foundation models. However, due to their particular training schemes and their design focus on broadening practical application scenarios, these models are typically tailored for single-image conditioning. 

\textbf{Multi-View 3D Generation Pipeline}
Several approaches have explored 3D generation with multi-view inputs.
Models such as TripoSG~\citep{li2025triposg} and Hunyuan3D~\citep{hunyuan3d2025hunyuan3d} are optimized with multiple views of the object during the training stage. Some works~\citep{feng2025arm,cheng2025mvpaint,wen2025ouroboros3d,gao2025charactershot} utilize the pre-trained multi-view diffusion models for more input images, aiding the \ryn{single-image generation} task. However, the above multi-image input pipelines are restricted to consistent views of the same object for the fixed model parameters. Some voxel-based generation methods like TRELLIS~\citep{xiang2024structured}, TRELLIS.2~\citep{xiang2025native} and Direct3D-S2~\citep{wu2025direct3d} adopt a simple strategy that switches the conditioning image at each denoising step to fuse information from multiple views. However, this often causes error accumulation and degrades the final 3D quality. In contrast to these methods, we propose a training-free pipeline that can handle diverse multi-image 3D tasks and robustly operate on inputs with viewpoint and appearance inconsistencies.

\textbf{3D Generation from Unconstrained Inputs}
\ryn{In 3D reconstruction, many} works~\citep{xu2024wild,martin2021nerf,kulhanek2024wildgaussians,sabour2025spotlesssplats,lin2025decoupling,go2025splatflow,qu2024lush} have investigated scene reconstruction from wild, unconstrained sets of input images. Yet, in the context of 3D generation, such methods can only recover the portions of an object that are directly observed and are unable to produce complete 3D models.
Recently, some works \ryn{investigate} 3D tasks with unconstrained images by interfering the 3D generation process, such as editing~\citep{qi2024tailor3d,jin2025fuse3d,zheng2025splatpainter}, style-guided generation~\citep{wang2024themestation,qu2025stylesculptor}, morphing~\citep{yin2025wukong, yang2025textured} and scene generation~\citep{huang2025midi,tan2021scenegen}. \ryn{However, these models are typically designed for specific tasks. To the best of our knowledge, no works have proposed a 3D generation pipeline that can process multiple unconstrained inputs of} the same object. Unlike prior works that focus on specific tasks like editing or morphing, UMI3D provides a general-purpose routing mechanism that generalizes single-image foundation models to arbitrary multi-image configurations.

\begin{figure*}
\vspace{-2mm}
  \includegraphics[width=\textwidth]{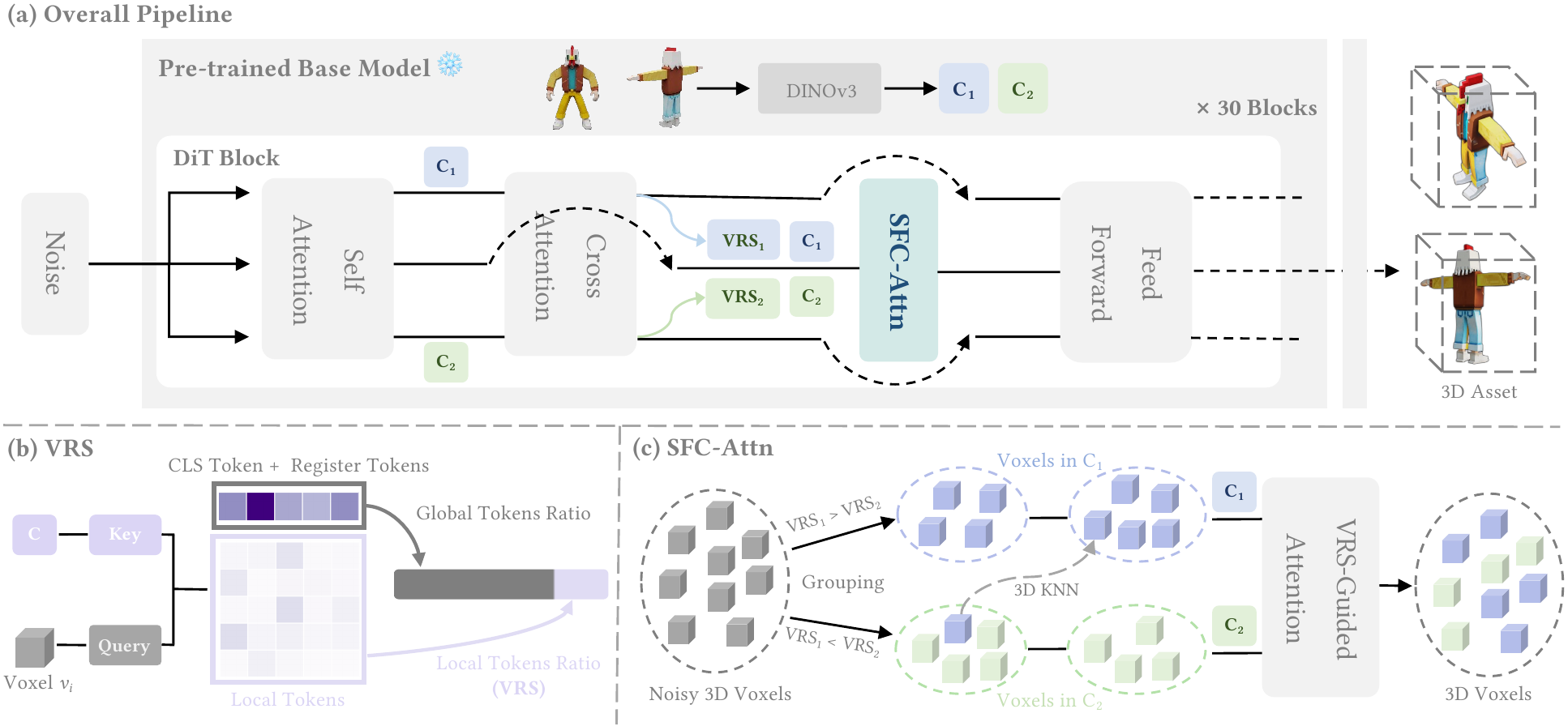}
  \vspace{-6mm}
  \caption{Overview of UMI3D. (a) Overall pipeline. 
(b) Voxel Reference Score (VRS). 
(c) Simultaneous Focus Cross-Attention (SFC-Attn). 
}
  \label{fig:pipeline}
  \vspace{-2mm}
\end{figure*}

\section{Method}
In this section, we present the architecture and underlying mechanism of UMI3D for 3D generation under multi-image inputs. We first give an overview of our pipeline (Sec.\ref{overview}), 
then introduce 
the \emph{Voxel Reference Score} (VRS, Sec.~\ref{sec:vrs}), a 
model-intrinsic metric that quantifies voxel--image affinity without any 
external supervision, and the \emph{Simultaneous Focus Cross-Attention} 
(SFC-Attn, Sec.~\ref{SFCA}) module, which leverages VRS to perform 
voxel-wise routing within each denoising step. The preliminary technique of our method is provided in the Appendix.

\subsection{Overview}
\label{overview}
The overall pipeline of UMI3D is illustrated in Fig.~\ref{fig:pipeline}(a). Given a set of unconstrained input images \( \{I_k\} \) of the same object, UMI3D operates on a pre-trained 3D generative model that takes Gaussian noise as input and progressively denoises it into a 3D asset in a zero-shot manner under multi-image conditioning.

We first introduce a Voxel Reference Score (VRS) metric (Sec.~\ref{sec:vrs}), as shown in Fig.~\ref{fig:pipeline}(b).
Starting from the baseline cross-attention (cross-attn) between 3D voxel tokens and 2D image features, VRS analyzes the attention maps of each voxel $v_i$ over different conditioning images and computes \ryn{a per-image score that indicates} how strongly the voxel is supported by each view.
These VRS \ryn{scores} are then used to partition the 3D voxels into subsets preferentially associated with different input images.
On top of VRS, we design the Simultaneous Focus Cross-Attention (SFC-Attn) module (Sec.~\ref{SFCA}), as depicted in Fig.~\ref{fig:pipeline}(c).
All input images are encoded by a DINOv3 encoder into multi-view features $\{C_k\}$.
Guided by the voxel-image assignments from VRS, SFC-Attn divides the 3D voxels into groups and performs cross-attn between each voxel group and its corresponding image features within the same DiT block.
The outputs of different groups are then merged and propagated through the following network.
\qzf{UMI3D does not introduce additional trainable parameters; it only modifies the inference-time attention computation through VRS estimation and SFC-Attn routing.}

\subsection{Voxel Reference Score (VRS)}
\label{sec:vrs}
For each conditioning image \(I_i\), we extract features using a frozen DINO image encoder, obtaining a token sequence:
\begin{equation}
    C_k = [\, \mathbf{t}^{\text{cls}},\, \mathbf{t}^{\text{reg}}_1, \dots, \mathbf{t}^{\text{reg}}_{N_r},\, \mathbf{t}^{\text{loc}}_1, \dots, \mathbf{t}^{\text{loc}}_{N_l} \,],
\end{equation}
where $\mathbf{t}^{\text{cls}}$ is the classification token, $\{\mathbf{t}^{\text{reg}}_{n_r}\}_{n_r=1}^{N_r}$ are the register tokens, and $\{\mathbf{t}^{\text{loc}}_{n_l}\}_{n_l=1}^{N_l}$ are local tokens corresponding to image patches.
We collectively refer to \(\mathbf{t}^{\text{cls}}\) and \(\mathbf{t}^{\text{reg}}\) as \emph{global tokens},
    $t^{\text{global}} = [\mathbf{t}^{\text{cls}}, \mathbf{t}^{\text{reg}}],$
as they capture rich global information about the entire image. In contrast, the local tokens are spatially aligned with image patches and primarily encode local \ryn{appearances} and fine-grained details.

\begin{figure}[t]
  \centering
  \vspace{-3mm}
  \begin{minipage}[t]{0.49\textwidth}
    \centering
    \includegraphics[width=\linewidth]{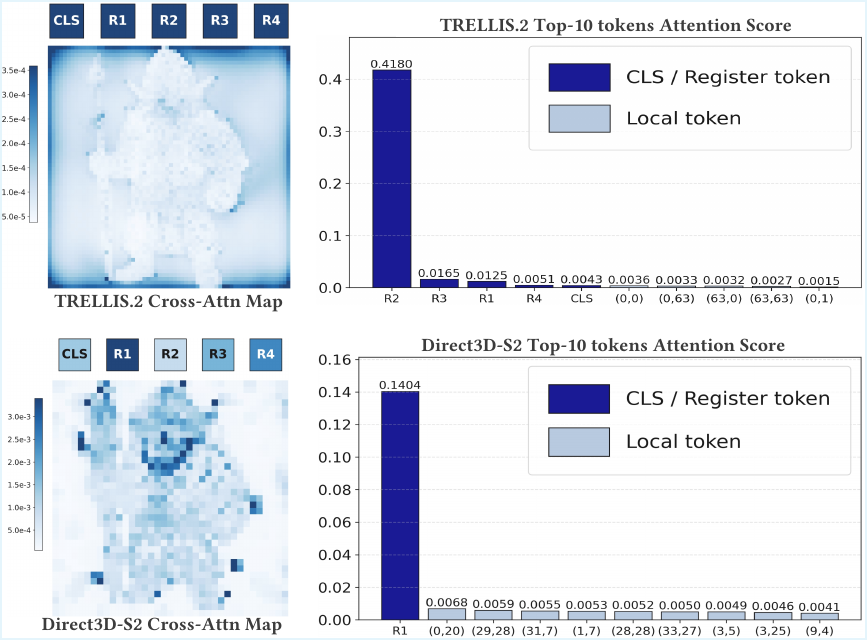}
    \caption{Insight 1 of the VRS metric.}
    \label{fig:vrs1}
  \end{minipage}
  \hfill 
  \begin{minipage}[t]{0.49\textwidth}
    \centering
    \includegraphics[width=\linewidth]{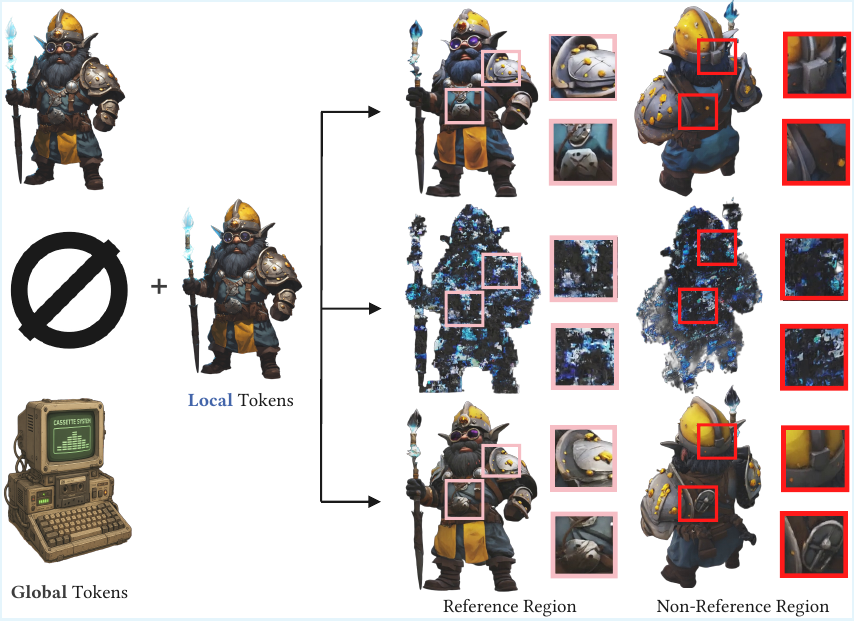}
    \caption{Insight 2 of the VRS metric. }
    \label{fig:vrs2}
  \end{minipage}
\end{figure}



By analyzing these cross-attn layers, we obtain two key insights that motivate our Voxel Reference Score (VRS):
\textbf{(1) Global tokens dominate the attention budget.}
As shown in Fig.~\ref{fig:vrs1} and Tab.~\ref{tab:insight_validation}, 
although global tokens constitute only a tiny fraction of the condition token sequence, voxel queries consistently allocate the majority of their attention mass to them rather than to the numerous local tokens. This disproportionate allocation indicates that the backbone primarily consults global tokens to reason about the object's overall structure and appearance.
\textbf{(2) Voxels lacking direct visual support fall back to global tokens.} 
Intuitively, a voxel that is well grounded in image $I_k$ should rely more on the image's patch-aligned local tokens, which provide spatially specific evidence; conversely, a voxel without direct visual support in $I_k$ has no localized evidence to attend to and instead falls back to the image-level semantics carried by global tokens.
In the Fig.~\ref{fig:vrs2}, when we replace the global tokens of the current view with those from \ryn{another arbitrary image} while keeping all local tokens fixed, the visible regions of the current view are only mildly affected, whereas the invisible regions change much more significantly.
These observations suggest that global tokens implicitly encode how well each image explains different parts of the 3D space. \ryn{This motivates us to explicitly quantify how strongly each voxel} is supported by each conditioning image.

\ryn{Specifically}, for each voxel $v_i$ and each input image $I_k$, we first obtain the corresponding query and key/value tokens from the cross-attn layer by Eq.~\ref{qkv}.
The attention map of voxel $v_i$ over the tokens of image $I_k$ is then computed as $\mathrm{AttnMap}_{i,k} = \mathrm{softmax}\bigl(\frac{\mathbf{Q}_i \mathbf{K}_k^\top}{\sqrt{d_q}}\bigr)$, where \(d_q\) denotes the feature dimension.
\(\mathrm{AttnMap}_{i,k} \in \mathbb{R}^{L_k}\) stores the attention weight from voxel \(v_i\) to each token of \(I_k\). Based on \(\mathrm{AttnMap}_{i,k}\), we define the VRS value as the total attention assigned to the local tokens of image \(I_k\), which is computed by subtracting the attention mass allocated to the global tokens:
\begin{equation}
\mathrm{VRS}_{i,k} 
\;=\; \sum_{t \in t_{\text{local}}} \mathrm{AttnMap}_{i,k}[t]
\;=\; 1 - \sum_{t \in t_{\text{global}}} \mathrm{AttnMap}_{i,k}[t].
\label{eq:global_score}
\end{equation}
Here, \(\mathrm{VRS}_{i,k}\) measures how strongly voxel \(v_i\) is supported by the local evidence provided by image \(I_k\) \ryn{-- a higher value indicates} that the voxel is well grounded in that image, while \ryn{a lower value indicates} fallback to global semantics. By comparing \(\mathrm{VRS}_{i,k}\) across different single-image conditioned generation branches \(k\), we can determine which reference image provides the highest-confidence support for voxel \(v_i\). We refer the reader to Sec.~\ref{exp_insight} for empirical validation of the proposed VRS metric.

\subsection{Simultaneous Focus Cross-Attention \ryn{(SFC-Attn)} Module}
\label{SFCA}
We now describe how the proposed 
\ryn{SFC-Attn} module leverages the Voxel Reference Score (VRS) to enable simultaneous yet selective conditioning on multiple input images within each DiT block of the pre-trained 3D backbone, as illustrated in Fig.~\ref{fig:pipeline}(a,c).

\vspace{1ex}

\noindent \ryn{\textbf{\textit{VRS Estimation Branches.}}}
Given the input images \(\{I_k\}_{k=1}^{K}\) and their corresponding DINOv3 token sequences \(\{C_k\}_{k=1}^{K}\), we first run multiple standard 3D conditioning branches, each conditioned on a \emph{single} image.
Concretely, for each image \(I_k\), we feed its tokens \(C_k\) into the original cross-attn layer of the frozen base model and obtain the 
\qzf{\(\mathrm{VRS}_{i,k}\)}
for every voxel \(v_i\), which measures how strongly voxel \(v_i\) is supported by image \(I_k\).
This stage consists of multiple \emph{parallel} cross-attn passes, each identical to the original single-image conditioning of the base model, and serves only to estimate voxel-image affinities without modifying any backbone parameters. 
Although each branch produces branch-specific queries under its own conditioning image, the resulting VRS values remain comparable across branches: all branches share the same denoising step, DiT block, input voxel \qzf{positions}, and backbone parameters, differing only in the conditioning image.
Therefore, VRS provides a consistent basis for comparing the relative support of different reference images for the same voxel. The validation experiments can be found in Sec.~\ref{exp_insight}.

\vspace{1ex}

\noindent \ryn{\textbf{\textit{Voxel Grouping by VRS.}}}
For each voxel \(v_i\), we define its dominant reference image as:
\begin{equation}
    \kappa(i) = \arg\max_{k} \, \mathrm{VRS}_{i,k},
\end{equation}
and construct $K$ disjoint voxel groups:
\begin{equation}
    \mathcal{G}_k = \{\, v_i \mid \kappa(i) = k \,\}, \quad k = 1,\dots,K.
\end{equation}
Thus, \(\mathcal{G}_k\) contains the voxels whose 3D \ryn{contents are} most strongly supported by image \(I_k\) according to VRS.
This produces a hard assignment that routes each voxel to a single conditioning image.

In 3D space, neighboring voxels typically belong to the same underlying surface region and therefore tend to be supported by the same conditioning image.
To enforce such spatial coherence, we further refine the coarse VRS-based grouping with a 3D \(k\)-nearest-neighbor (KNN) scheme.
Let the coordinate of voxel $v_i$ be \(\mathbf{p}_i = (x_i, y_i, z_i)\).
For each voxel $v_i$, we consider its $K_{\text{nn}}$ nearest neighbors in terms of 3D Euclidean distance $d(i,j)$,
and denote their index set by $\mathcal{N}_K(i)$.
We refine the routing decision of each voxel by taking the mode of the dominant views within its 3D KNN neighborhood:
\begin{equation}
    \tilde{\kappa}(i)
    = \mathrm{Mode}\big(\{\kappa(j)\}_{j \in \mathcal{N}_K(i)}\big),
\end{equation}
and update the voxel groups accordingly, as:
\begin{equation}
    \tilde{\mathcal{G}}_k
    = \{\, v_i \mid \tilde{\kappa}(i) = k \,\}, \quad k = 1,\dots,K.
\end{equation}

This refinement brings two benefits:
(i) it mitigates the influence of noisy or outlier VRS responses at individual voxels, leading to \ryn{a} more reliable view assignment; and
\qzf{(ii) it encourages ambiguous voxels with weak direct evidence to follow the dominant reference of neighboring, better-supported voxels, thereby reducing feature discontinuities in ambiguous regions.}

\vspace{1ex}

\noindent \ryn{\textbf{\textit{VRS-guided Attention.}}}
After obtaining the voxel groups $\{\tilde{\mathcal{G}}_k\}$, SFC-Attn performs view-specific cross-attn for each group.
Let $\mathbf{Q}_k \in \mathbb{R}^{|\tilde{\mathcal{G}}_k| \times d}$ be the sub-matrix of queries corresponding to voxels in $\tilde{\mathcal{G}}_k$.
For each image $I_k$, we compute keys $\mathbf{K}_k$ and values $\mathbf{V}_k$ from $C_k$, \ryn{and then obtain} the SFC-Attn output for voxels in $\tilde{\mathcal{G}}_k$ as:
\begin{equation}
    \mathbf{Z}_k
    = \mathrm{Softmax}\!\left(
        \frac{\mathbf{Q}_k \mathbf{K}_k^\top}{\sqrt{d_q}}
    \right)\mathbf{V}_k,
    \qquad k = 1,\dots,K,
    \label{eq:sfc_attn_group}
\end{equation}
where \(\mathbf{Z}_k \in \mathbb{R}^{|\tilde{\mathcal{G}}_k| \times d}\) contains the updated features of all voxels in group \(\tilde{\mathcal{G}}_k\), extracted only from the conditioning image \(I_k\).
Finally, the group-wise outputs \(\{\mathbf{Z}_k\}_{k=1}^{K}\) are scattered back to their original voxel indices to recover the full voxel feature matrix \(\mathbf{Z}\). Since the refined groups $\{\tilde{\mathcal{G}}_k\}$ are disjoint by construction, this merge is unambiguous.

In this way, all input images are \emph{simultaneously} available within the same denoising step, while each voxel \emph{focuses} on the single image that best explains it.
The entire mechanism is training-free and keeps the pre-trained 3D backbone intact, \ryn{and} yet substantially improves multi-image 3D generation by replacing the implicit, oscillating conditioning of naive baselines with a structured, VRS-guided routing of 
multi-view evidence.

\begin{figure*}
\vspace{-3mm}
  \includegraphics[width=\textwidth]{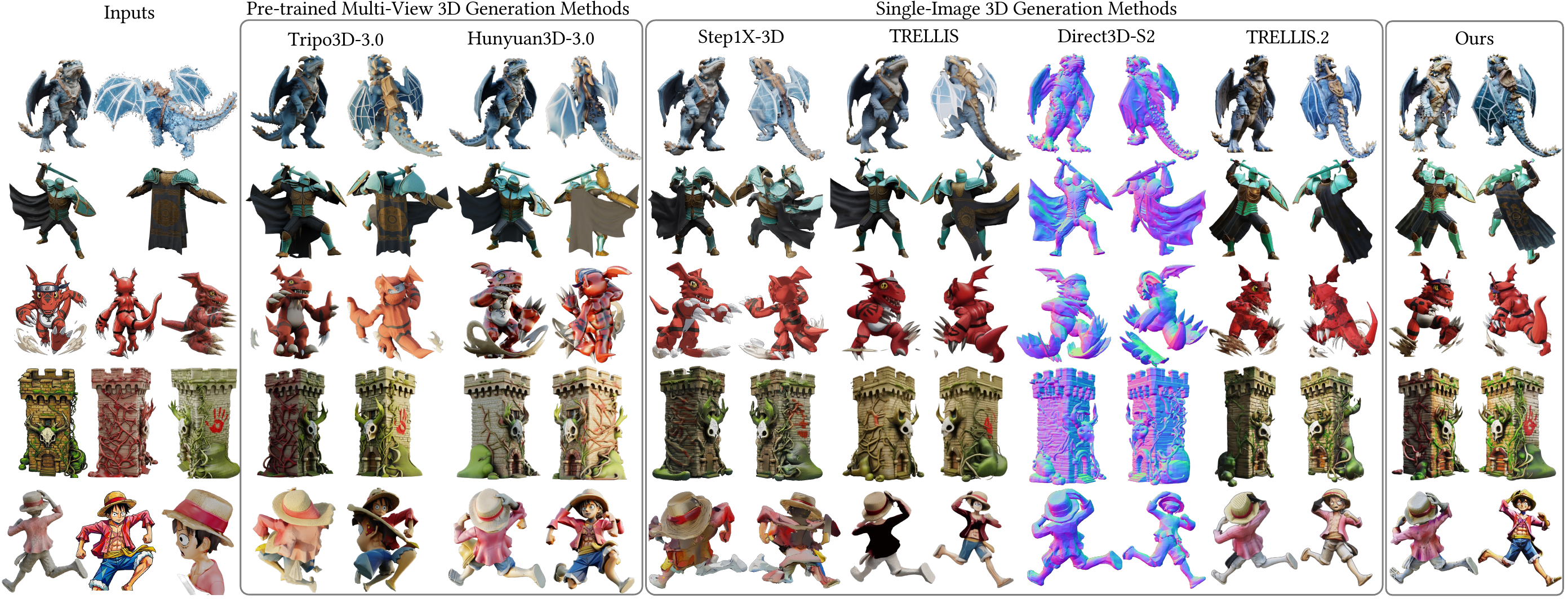}
  \caption{Qualitative comparison to multi-view 3D generation model, Tripo-3D~\citep{li2025triposg}, Hunyuan3D~\citep{hunyuan3d2025hunyuan3d} and single-view 3D generation pipeline Step1X-3D~\citep{li2025step1x}, TRELLIS~\citep{xiang2024structured}, Direct3D-S2~\citep{wu2025direct3d}, and TRELLIS.2~\citep{xiang2025native}.
  }
  \label{fig:comparison}
\end{figure*}

\section{Experiments}
In this section, 
We first conduct qualitative and quantitative comparisons, together with a user study, against existing state-of-the-art methods. We \ryn{further} perform ablation studies to validate the effectiveness of each proposed component\ryn{, and} validate the two insights underlying the VRS metric in Sec.~\ref{sec:vrs}. Finally, we showcase additional applications enabled by UMI3D. More results and implementation details can be found in \textbf{Appendix}.

\subsection{Comparisons with State-of-the-Art Methods}
\subsubsection{Methods Compared.}
To the best of our knowledge, UMI3D is the first approach that enables unconstrained multi-image input for 3D generation. Since no prior work directly addresses this task, we compare UMI3D with the approaches mentioned in Sec.~\ref{intro}: (1) pre-trained multi-view 3D generation approaches and (2) naive multi-image extensions of single-image 3D generation methods.
For \ryn{(1), we select} Hunyuan3D 3.0~\citep{hunyuan3d2025hunyuan3d} and Tripo3D 3.0~\citep{li2025triposg} for testing. \ryn{Note that these} are commercial models, and we use their official website interfaces for inference. For \ryn{(2)}, we compare with Step1X-3D~\citep{li2025step1x}, TRELLIS~\citep{xiang2024structured}, Direct3D-S2~\citep{wu2025direct3d}, and TRELLIS.2~\citep{xiang2025native}. For Direct3D-S2, TRELLIS and TRELLIS.2, only the anchor image is used as the condition in the sparse structure generation stage, following their default single-image settings.  Since Direct3D-S2 cannot generate textures, we use the normal maps \ryn{that it produces} as a substitute in the visualizations.

\subsubsection{Qualitative Results.}
As shown in Fig.~\ref{fig:comparison} and~\ref{fig:comparison_all}, we present extensive comparisons with existing \ryn{SOTA} 3D generation methods.
(1) For \emph{Pre-trained multi-view generation models}, Tripo3D and Hunyuan3D exhibit strong single-image 3D generation capabilities, but struggle to handle unconstrained multi-image inputs with heavy occlusions, large pose changes, or significant semantic differences. Since these models are trained with multi-view image sets that are strictly consistent in viewpoint and appearance, extending them to such unconstrained settings often leads to confused, misaligned, or entangled 3D results.
(2) For \emph{naive generation pipelines with single-image based models},  they are limited by the fact that only one image can be used as the condition at each step, which prevents the model from leveraging global information across all inputs and tends to cause problems like overfitting to a single condition, and missing, over-smooth or inconsistent geometry and textures. In contrast, UMI3D attends to all conditioning images simultaneously at each denoising step while selectively extracting the most  informative evidence for each 3D region, yielding coherent and  high-quality 3D generation under unconstrained multi-image inputs.

\begin{table}[]
\vspace{-3mm}
\caption{\textbf{Quantitative evaluation.} All results are tallied across 40 different generation results.}
\label{tab:quan}
\centering
\resizebox{\columnwidth}{!}{%
\begin{tabular}{c|cccc}
\hline
\textbf{Metric}        & \textbf{CLIP $\uparrow$} & \textbf{DINO $\uparrow$} & \textbf{LPIPS $\downarrow$} & \textbf{User Study (\%) $\uparrow$} \\ \hline
Tripo3D-3.0~\citep{li2025triposg}   &  85.97    & 71.76     &  0.3361     &  4.98               \\
Hunyuan3D-3.0~\citep{hunyuan3d2025hunyuan3d} & 85.41     &   69.67   &  0.3408     &  7.75               \\ \hline
Step1X-3D~\citep{li2025step1x}     & 85.52     &  69.01    &  0.3433     & 5.68                \\
TRELLIS~\citep{xiang2024structured}       &  85.60    & 70.44     &  0.3429     &  5.16               \\
TRELLIS.2~\citep{xiang2025native}     &  85.95    &  72.99    &  0.3413       & 15.86                \\ \hline
\textbf{UMI3D (Ours)}  & \textbf{86.42}     &  \textbf{74.62}    &  \textbf{0.3356}       & \textbf{60.52}                \\ \hline
\end{tabular}%
}
\end{table}

\subsubsection{Quantitative Results.}


As shown in Tab.~\ref{tab:quan}, our method achieves the best performance on all four metrics. The CLIP score mainly reflects high-level semantic consistency. Since all input images correspond to the same semantic object, the differences among the methods are relatively small. Nevertheless, UMI3D still outperforms the second-best method, Tripo3D 3.0, by 0.45 points. DINO is more sensitive to local details, and UMI3D improves the DINO score by 1.63 points over TRELLIS.2, achieving the best performance for unconstrained multi-image 3D generation. 
The user study provides the most discriminative signal: \textbf{60.52\%} of UMI3D's results are rated as the best across the six baselines, more than $3.8{\times}$ higher than the runner-up, indicating that UMI3D produces better results for both reference and unseen viewpoints.

\subsection{Ablation Study}

\begin{figure*}[t]
  \includegraphics[width=\textwidth]{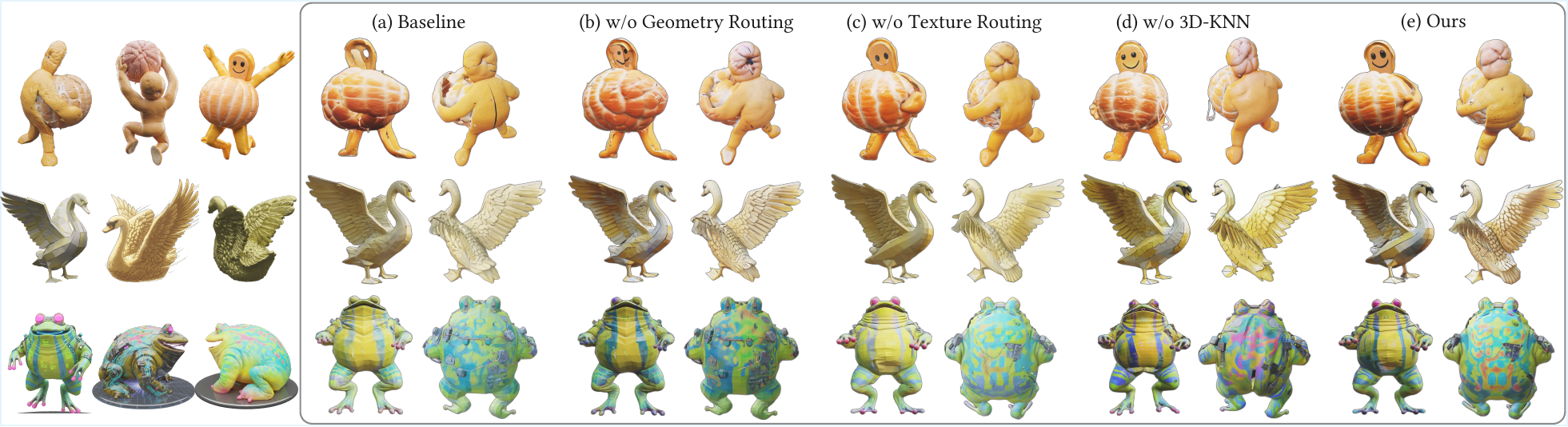}
  \caption{Overall ablation study on UMI3D. (a) The Baseline Model (TRELLIS.2).
(b) w/o Geometry Routing.
(c) w/o Texture Routing.
(d) w/o 3D-KNN. 
(e) \ryn{Our Full UMI3D Model}.
}
  \label{fig:ablation}
\end{figure*}

To evaluate the effectiveness of each component, we conduct an ablation study as shown in Fig.~\ref{fig:ablation}.
\textbf{Effectiveness of SFC-Attn in Geometry Generation.}
As shown in (b) and (e), removing SFC-Attn from the geometry generation stage of UMI3D leads to errors in the structural information of the generated results (such as the face of the orange character), and inaccuracies in the structural information can also make it difficult to align the texture information, thereby affecting the results across multiple stages.
\textbf{Effectiveness of SFC-Attn in Texture Generation.} The absence of SFC-Attn during the texture generation stage can lead to issues such as overly smooth textures (e.g., the color discrepancy at the front of the frog) and loss of detail (e.g., the eyes of the swan).
\textbf{Effectiveness of the 3D-KNN operation.} Comparing (d) and (e), the absence of the 3D-KNN mechanism leads to some local discrepancies. \ryn{Incorporating} KNN processing improves routing accuracy, thereby producing high-quality 3D generation results.


\begin{table}[htbp]
\centering
\begin{minipage}{0.58\textwidth}
    \centering
    \caption{The attention score of different types of tokens.}
    \resizebox{\linewidth}{!}{%
    \begin{tabular}{c|ccc}
    \hline
    \textbf{Ratio (Per-Token Ratio)}       & \textbf{CLS Token} & \textbf{Register Tokens} & \textbf{Local Tokens} \\ \hline
    TRELLIS (Stage-2)     &  0.0014 (0.0014)         &   0.8105 (0.2026)              &  0.1881 (0.0001)            \\
    Direct3D-S2 (Stage-2) &  0.0022 (0.0022)         &  0.1541 (0.0385)               &  0.8437 (0.0006)            \\
    TRELLIS.2 (Stage-2)   &  0.0054 (0.0054)         & 0.4835 (0.1209)                & 0.5111 (0.0001)             \\ 
    TRELLIS.2 (Stage-3)   &  0.0050 (0.0050)         & 0.5335 (0.1334)                &  0.4615 (0.0001)            \\ \hline
    \end{tabular}
    }
    \label{tab:insight_validation}
\end{minipage}
\hfill 
\begin{minipage}{0.4\textwidth}
    \centering
    \caption{Attention Routing Strategies.}
    \resizebox{\linewidth}{!}{%
    \begin{tabular}{c|ccc}
    \hline
    \textbf{Metric}        & \textbf{CLIP $\uparrow$} & \textbf{DINO $\uparrow$} & \textbf{LPIPS $\downarrow$}  \\ \hline
    Baseline   &  85.95    &  72.99    &  0.3413                    \\
    Random CA Routing & 85.61     & 72.36     &  0.3423                    \\ 
    Opposite VRS Routing     & 84.42     &  69.63    &   0.3439                   \\
    VRS Routing (Ours)     & \textbf{86.42}     &  \textbf{74.62}    &  \textbf{0.3356}                    \\  \hline
    \end{tabular}
    }
    \label{tab:insight_routing}
\end{minipage}
\end{table}

\subsection{Insight Validation} \label{exp_insight}
\ryn{Here}, we experimentally validate the two insights proposed in Sec.~\ref{sec:vrs}, providing empirical support for both the VRS metric and SFC-Attn.

\begin{figure*}[h]
\vspace{-4mm}
  \includegraphics[width=\textwidth]{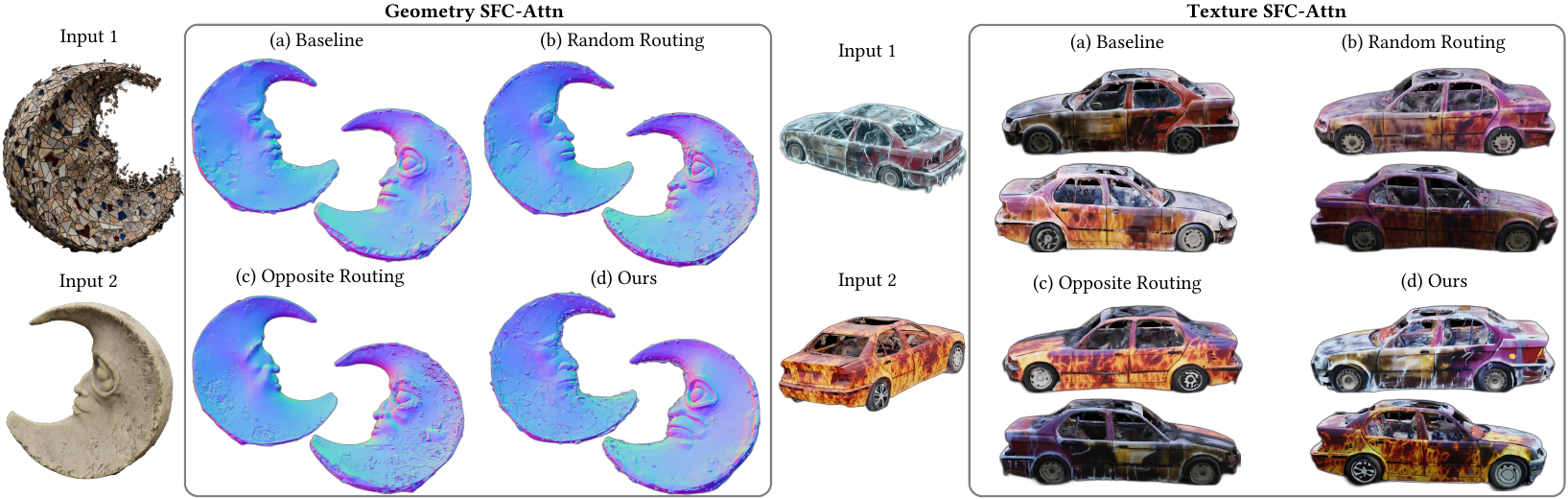}
  \caption{Insight validation of the SFC-Attn module.
(a) Baseline.
(b) Random Routing: Voxels are randomly routed to condition images.
(c) Opposite Routing: Voxels are routed to the condition image with the lowest VRS score.
(d) Ours: Voxels are routed with the highest VRS score.
}
\vspace{-2mm}
  \label{fig:insight_valid}
\end{figure*}

To validate \textbf{Insight 1}, we analyze the attention scores assigned to different types of tokens in the condition image during single-image generation with TRELLIS, Direct3D-S2, and TRELLIS.2.
\qzf{This analysis is conducted on 41 individual images from the official TRELLIS repository, which are separate from our 40-case multi-image benchmark.}
For each model, we average the attention scores over voxels and attention heads, and then accumulate them across all denoising steps. The resulting per-token scores, together with the number of tokens of each type, are reported in Tab.~\ref{tab:insight_validation}.

The results reveal that, when extracting information from a 2D image, these 3D generation models rely heavily on register tokens: in TRELLIS, the 4 register tokens alone account for 81\% of the total attention, whereas the remaining 1{,}369 local tokens collectively contribute only about 19\%. Although the CLS token receives a smaller share than the register tokens, its individual contribution is still substantially higher than that of any single local token. Based on this observation, we group the CLS token together with the register tokens as global tokens when computing the VRS metric.

To validate \textbf{Insight 2}, we present ablation results on UMI3D in Fig.~\ref{fig:insight_valid} and Tab.~\ref{tab:insight_routing}.
As shown in \ryn{Fig.~\ref{fig:insight_valid}(a,b)}, for both geometry and texture generation, random routing without the VRS metric degrades the baseline performance, since it \ryn{completely} ignores the alignment between 3D voxels and the conditional images. Consistently, Tab.~\ref{tab:insight_routing} shows that Random Cross-Attn Routing performs slightly worse than the baseline.

Fig.~\ref{fig:insight_valid}(c) further shows that when each voxel selects the image with the smallest VRS value as its condition, the resulting geometry and texture become inverted with respect to the evidence provided by the inputs. For example, the rough and smooth surface cues of the moon are swapped, and the flame-like and frosted appearances of the car are assigned in reverse to what the inputs indicate. By forcing each voxel to attend to the least relevant image, this opposite-VRS routing compels the model to extract information from references that bear minimal correspondence to the queried voxel, which in turn confirms that VRS effectively captures the underlying 2D--3D correspondence. The first and third rows in Tab.~\ref{tab:insight_routing} corroborate this finding, where Opposite-VRS routing leads to a significant performance drop, as each voxel ends up selecting the image with the least relevant reference information and the generated results deviate substantially from the inputs.

In contrast, as shown in Fig.~\ref{fig:insight_valid}(d), routing based on the correct VRS metric enables the model to accurately fuse complementary information across views with large appearance differences, yielding a faithful 3D asset.

\section{\ryn{Applications}}
\begin{figure*}[h]
  \vspace{-5mm}
  \centering
  \includegraphics[width=0.97\textwidth]{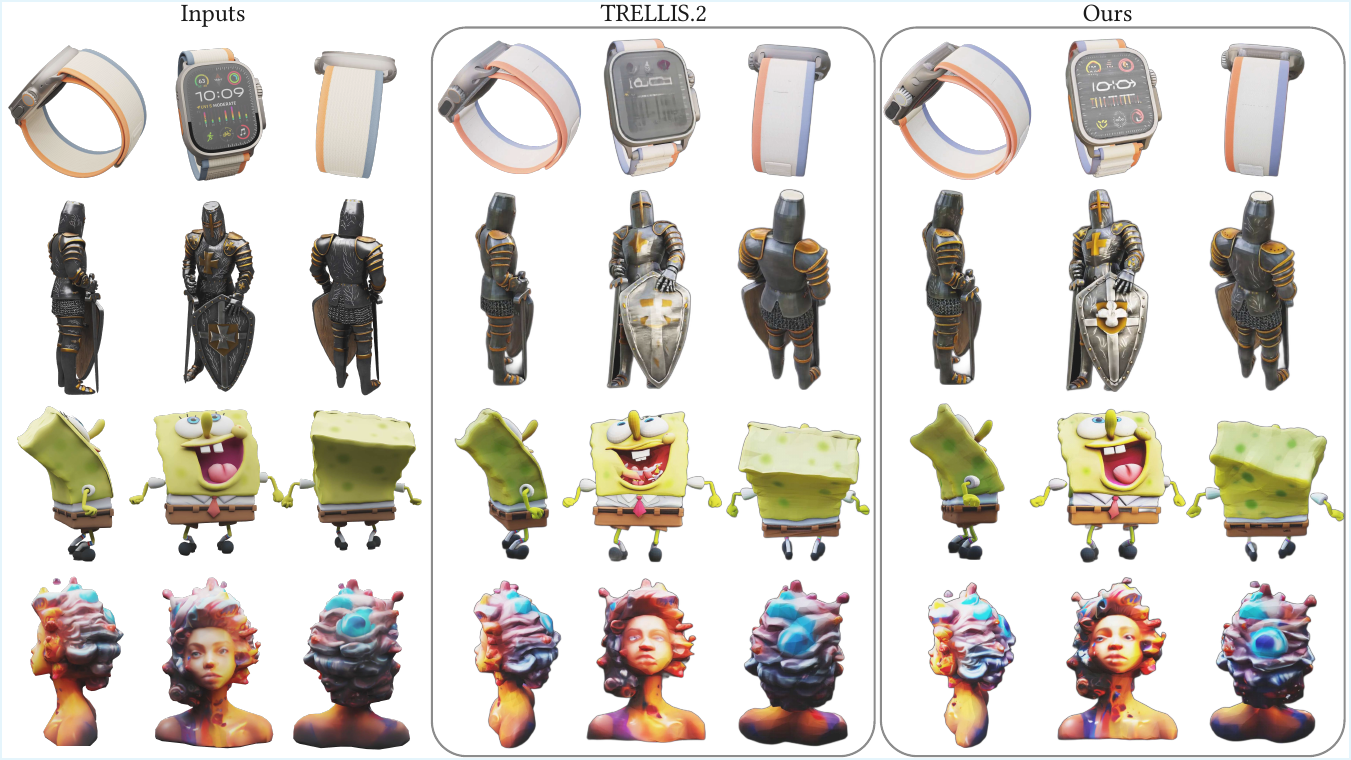}
  \caption{The multi-view 3D generation enhancement results of UMI3D.}
  \label{fig:multi-view}
\end{figure*}

\begin{figure*}[h]
    \centering
  \includegraphics[width=0.97\textwidth]{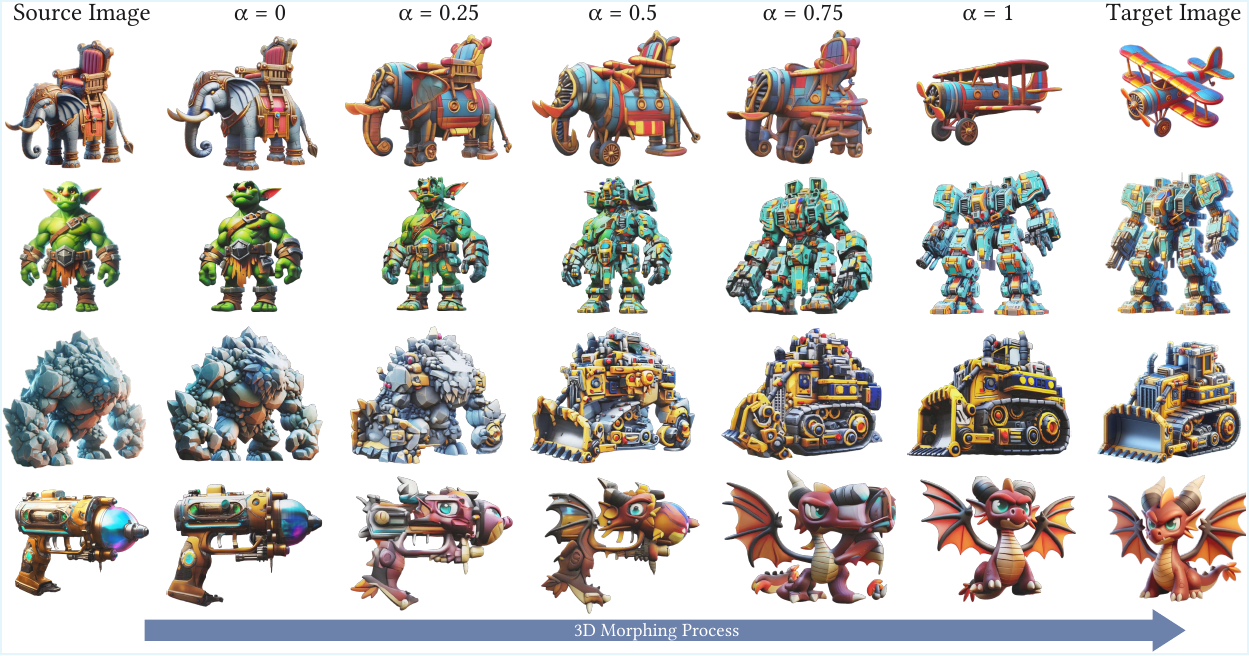}
  \caption{The 3D morphing results of UMI3D. $\alpha$ controls the 3D morphing progress.
  }
  \label{fig:morphing}
\end{figure*}

Beyond unconstrained multi-image 3D generation, UMI3D also enables three practical applications: multi-view 3D generation enhancement, 3D morphing, and single-view local enhancement, as shown in Figs.~\ref{fig:multi-view},  \ref{fig:morphing}, and \ref{fig:single-view}. These examples further highlight the versatility of UMI3D. Detailed settings and analysis are deferred to the appendix.



\section{Conclusion}

This paper introduces \textit{UMI3D}, a training-free multi-image 3D generation framework that supports robust 3D synthesis from unconstrained image collections. UMI3D incorporates a novel SFC-Attn module, which enables voxel-wise simultaneous and selective attention over multiple conditioning images to effectively integrate complementary visual information during generation. To address the ambiguity of voxel-image association in existing 3D generation models, UMI3D leverages the intrinsic behavior of cross-attention layers to derive the Voxel Reference Score (VRS), allowing the model to automatically establish reliable voxel-to-image correspondences without requiring external priors or auxiliary models. Finally, by reformulating multi-image conditioning as a dynamic routing process in the 3D latent space, UMI3D unlocks the multi-image capability of single-image 3D foundation models and achieves robust and coherent 3D generation in diverse real-world settings.



\paragraph{Limitations.}

As shown in Fig.~\ref{fig:failure}, UMI3D may fail to generate accurate 3D assets when the target object exhibits strong symmetry, such as \ryn{a sphere or an} object with multiple highly similar faces. This limitation arises because UMI3D relies on the implicit 2D-3D alignment priors embedded in the pre-trained large-scale 3D generative model to perform unconstrained generation. For such highly \ryn{symmetrical} objects, however, these implicit alignment cues become less reliable, which in turn degrades the effectiveness of VRS-based cross-attention routing and leads to unstable or chaotic generation results. 
We will explore this problem in our future works.

\bibliography{iclr2026_conference}

\begin{thebibliography}{53}
\providecommand{\natexlab}[1]{#1}
\providecommand{\url}[1]{\texttt{#1}}
\expandafter\ifx\csname urlstyle\endcsname\relax
  \providecommand{\doi}[1]{doi: #1}\else
  \providecommand{\doi}{doi: \begingroup \urlstyle{rm}\Url}\fi

\bibitem[Chen et~al.(2023)Chen, Gu, Chen, Tian, Tu, Liu, and Su]{chen2023single}
Hansheng Chen, Jiatao Gu, Anpei Chen, Wei Tian, Zhuowen Tu, Lingjie Liu, and Hao Su.
\newblock Single-stage diffusion nerf: A unified approach to 3d generation and reconstruction.
\newblock In \emph{Proceedings of the IEEE/CVF International Conference on Computer Vision}, pp.\  2416--2425, 2023.

\bibitem[Chen et~al.(2025)Chen, Chu, Gleize, Liang, Sax, Tang, Wang, Guo, Hardin, Li, et~al.]{chen2025sam}
Xingyu Chen, Fu-Jen Chu, Pierre Gleize, Kevin~J Liang, Alexander Sax, Hao Tang, Weiyao Wang, Michelle Guo, Thibaut Hardin, Xiang Li, et~al.
\newblock Sam 3d: 3dfy anything in images.
\newblock \emph{arXiv preprint arXiv:2511.16624}, 2025.

\bibitem[Chen \& Zhang(2019)Chen and Zhang]{chen2019net}
Zhiqin Chen and Hao Zhang.
\newblock Learning implicit fields for generative shape modeling.
\newblock In \emph{Proceedings of the IEEE/CVF Conference on Computer Vision and Pattern Recognition}, pp.\  5939--5948, 2019.

\bibitem[Cheng et~al.(2025)Cheng, Mu, Zeng, Chen, Pang, Zhang, Wang, Fu, Yu, Liu, et~al.]{cheng2025mvpaint}
Wei Cheng, Juncheng Mu, Xianfang Zeng, Xin Chen, Anqi Pang, Chi Zhang, Zhibin Wang, Bin Fu, Gang Yu, Ziwei Liu, et~al.
\newblock Mvpaint: Synchronized multi-view diffusion for painting anything 3d.
\newblock In \emph{Proceedings of the Computer Vision and Pattern Recognition Conference}, pp.\  585--594, 2025.

\bibitem[Feng et~al.(2025)Feng, Yu, Bi, Shang, Gao, Wu, Zhou, Jiang, and Yang]{feng2025arm}
Xiang Feng, Chang Yu, Zoubin Bi, Yintong Shang, Feng Gao, Hongzhi Wu, Kun Zhou, Chenfanfu Jiang, and Yin Yang.
\newblock Arm: appearance reconstruction model for relightable 3d generation.
\newblock In \emph{Proceedings of the Computer Vision and Pattern Recognition Conference}, pp.\  21425--21437, 2025.

\bibitem[Gao et~al.(2025)Gao, Li, Liu, Zeng, Shen, Chen, Sun, and Zhao]{gao2025charactershot}
Junyao Gao, Jiaxing Li, Wenran Liu, Yanhong Zeng, Fei Shen, Kai Chen, Yanan Sun, and Cairong Zhao.
\newblock Charactershot: Controllable and consistent 4d character animation.
\newblock \emph{arXiv preprint arXiv:2508.07409}, 2025.

\bibitem[Go et~al.(2025)Go, Park, Jang, Kim, Kwon, and Kim]{go2025splatflow}
Hyojun Go, Byeongjun Park, Jiho Jang, Jin-Young Kim, Soonwoo Kwon, and Changick Kim.
\newblock Splatflow: Multi-view rectified flow model for 3d gaussian splatting synthesis.
\newblock In \emph{Proceedings of the Computer Vision and Pattern Recognition Conference}, pp.\  21524--21536, 2025.

\bibitem[Hertz et~al.(2022)Hertz, Perel, Giryes, Sorkine-Hornung, and Cohen-Or]{hertz2022spaghetti}
Amir Hertz, Or~Perel, Raja Giryes, Olga Sorkine-Hornung, and Daniel Cohen-Or.
\newblock Spaghetti: Editing implicit shapes through part aware generation.
\newblock \emph{ACM Transactions on Graphics (TOG)}, 41\penalty0 (4):\penalty0 1--20, 2022.

\bibitem[Ho et~al.(2020)Ho, Jain, and Abbeel]{ho2020denoising}
Jonathan Ho, Ajay Jain, and Pieter Abbeel.
\newblock Denoising diffusion probabilistic models.
\newblock \emph{Advances in Neural Information Processing Systems}, 33:\penalty0 6840--6851, 2020.

\bibitem[Huang et~al.(2025)Huang, Guo, An, Yang, Li, Zou, Liang, Liu, Cao, and Sheng]{huang2025midi}
Zehuan Huang, Yuan-Chen Guo, Xingqiao An, Yunhan Yang, Yangguang Li, Zi-Xin Zou, Ding Liang, Xihui Liu, Yan-Pei Cao, and Lu~Sheng.
\newblock Midi: Multi-instance diffusion for single image to 3d scene generation.
\newblock In \emph{Proceedings of the Computer Vision and Pattern Recognition Conference}, pp.\  23646--23657, 2025.

\bibitem[Hui et~al.(2022)Hui, Li, Hu, and Fu]{hui2022neural}
Ka-Hei Hui, Ruihui Li, Jingyu Hu, and Chi-Wing Fu.
\newblock Neural wavelet-domain diffusion for 3d shape generation.
\newblock In \emph{SIGGRAPH Asia 2022 Conference Papers}, pp.\  1--9, 2022.

\bibitem[Hui et~al.(2024)Hui, Sanghi, Rampini, Malekshan, Liu, Shayani, and Fu]{hui2024make}
Ka-Hei Hui, Aditya Sanghi, Arianna Rampini, Kamal~Rahimi Malekshan, Zhengzhe Liu, Hooman Shayani, and Chi-Wing Fu.
\newblock Make-a-shape: a ten-million-scale 3d shape model.
\newblock In \emph{Forty-first International Conference on Machine Learning}, 2024.

\bibitem[Jin et~al.(2025)Jin, Xie, Zheng, Wang, Bao, and Huo]{jin2025fuse3d}
Xuancheng Jin, Rengan Xie, Wenting Zheng, Rui Wang, Hujun Bao, and Yuchi Huo.
\newblock Fuse3d: Generating 3d assets controlled by multi-image fusion.
\newblock In \emph{Proceedings of the SIGGRAPH Asia 2025 Conference Papers}, pp.\  1--12, 2025.

\bibitem[Kulhanek et~al.(2024)Kulhanek, Peng, Kukelova, Pollefeys, and Sattler]{kulhanek2024wildgaussians}
Jonas Kulhanek, Songyou Peng, Zuzana Kukelova, Marc Pollefeys, and Torsten Sattler.
\newblock Wildgaussians: 3d gaussian splatting in the wild.
\newblock \emph{arXiv preprint arXiv:2407.08447}, 2024.

\bibitem[Li et~al.(2025{\natexlab{a}})Li, Zhang, Sun, Qi, Li, Cheng, Cai, Wu, Liu, Wang, et~al.]{li2025step1x}
Weiyu Li, Xuanyang Zhang, Zheng Sun, Di~Qi, Hao Li, Wei Cheng, Weiwei Cai, Shihao Wu, Jiarui Liu, Zihao Wang, et~al.
\newblock Step1x-3d: Towards high-fidelity and controllable generation of textured 3d assets.
\newblock \emph{arXiv:2505.07747}, 2025{\natexlab{a}}.

\bibitem[Li et~al.(2025{\natexlab{b}})Li, Zou, Liu, Wang, Liang, Yu, Liu, Guo, Liang, Ouyang, et~al.]{li2025triposg}
Yangguang Li, Zi-Xin Zou, Zexiang Liu, Dehu Wang, Yuan Liang, Zhipeng Yu, Xingchao Liu, Yuan-Chen Guo, Ding Liang, Wanli Ouyang, et~al.
\newblock Triposg: High-fidelity 3d shape synthesis using large-scale rectified flow models.
\newblock \emph{arXiv preprint arXiv:2502.06608}, 2025{\natexlab{b}}.

\bibitem[Li et~al.(2025{\natexlab{c}})Li, Wang, Zheng, Luo, and Wen]{li2025sparc3d}
Zhihao Li, Yufei Wang, Heliang Zheng, Yihao Luo, and Bihan Wen.
\newblock Sparc3d: Sparse representation and construction for high-resolution 3d shapes modeling.
\newblock \emph{arXiv preprint arXiv:2505.14521}, 2025{\natexlab{c}}.

\bibitem[Lin et~al.(2025)Lin, Li, Huang, Tang, Liu, Liu, Wu, Song, and Yang]{lin2025decoupling}
Jiaqi Lin, Zhihao Li, Binxiao Huang, Xiao Tang, Jianzhuang Liu, Shiyong Liu, Xiaofei Wu, Fenglong Song, and Wenming Yang.
\newblock Decoupling appearance variations with 3d consistent features in gaussian splatting.
\newblock In \emph{Proceedings of the AAAI Conference on Artificial Intelligence}, volume~39, pp.\  5236--5244, 2025.

\bibitem[Liu et~al.(2022)Liu, Gong, and Liu]{liu2022flow}
Xingchao Liu, Chengyue Gong, and Qiang Liu.
\newblock Flow straight and fast: Learning to generate and transfer data with rectified flow.
\newblock \emph{arXiv preprint arXiv:2209.03003}, 2022.

\bibitem[Luo \& Hu(2021)Luo and Hu]{luo2021diffusion}
Shitong Luo and Wei Hu.
\newblock Diffusion probabilistic models for 3d point cloud generation.
\newblock In \emph{Proceedings of the IEEE/CVF Conference on Computer Vision and Pattern Recognition}, pp.\  2837--2845, 2021.

\bibitem[Martin-Brualla et~al.(2021)Martin-Brualla, Radwan, Sajjadi, Barron, Dosovitskiy, and Duckworth]{martin2021nerf}
Ricardo Martin-Brualla, Noha Radwan, Mehdi~SM Sajjadi, Jonathan~T Barron, Alexey Dosovitskiy, and Daniel Duckworth.
\newblock Nerf in the wild: Neural radiance fields for unconstrained photo collections.
\newblock In \emph{Proceedings of the IEEE/CVF Conference on Computer Vision and Pattern Recognition}, pp.\  7210--7219, 2021.

\bibitem[M{\"u}ller et~al.(2023)M{\"u}ller, Siddiqui, Porzi, Bulo, Kontschieder, and Nie{\ss}ner]{muller2023diffrf}
Norman M{\"u}ller, Yawar Siddiqui, Lorenzo Porzi, Samuel~Rota Bulo, Peter Kontschieder, and Matthias Nie{\ss}ner.
\newblock Diffrf: Rendering-guided 3d radiance field diffusion.
\newblock In \emph{Proceedings of the IEEE/CVF Conference on Computer Vision and Pattern Recognition}, pp.\  4328--4338, 2023.

\bibitem[Nichol et~al.(2022)Nichol, Jun, Dhariwal, Mishkin, and Chen]{nichol2022point}
Alex Nichol, Heewoo Jun, Prafulla Dhariwal, Pamela Mishkin, and Mark Chen.
\newblock Point-e: A system for generating 3d point clouds from complex prompts.
\newblock \emph{arXiv preprint arXiv:2212.08751}, 2022.

\bibitem[Oquab et~al.(2023)Oquab, Darcet, Moutakanni, Vo, Szafraniec, Khalidov, Fernandez, Haziza, Massa, El-Nouby, Howes, Huang, Xu, Sharma, Li, Galuba, Rabbat, Assran, Ballas, Synnaeve, Misra, Jegou, Mairal, Labatut, Joulin, and Bojanowski]{oquab2023dinov2}
Maxime Oquab, Timothée Darcet, Theo Moutakanni, Huy~V. Vo, Marc Szafraniec, Vasil Khalidov, Pierre Fernandez, Daniel Haziza, Francisco Massa, Alaaeldin El-Nouby, Russell Howes, Po-Yao Huang, Hu~Xu, Vasu Sharma, Shang-Wen Li, Wojciech Galuba, Mike Rabbat, Mido Assran, Nicolas Ballas, Gabriel Synnaeve, Ishan Misra, Herve Jegou, Julien Mairal, Patrick Labatut, Armand Joulin, and Piotr Bojanowski.
\newblock Dinov2: Learning robust visual features without supervision, 2023.

\bibitem[Peebles \& Xie(2022)Peebles and Xie]{Peebles2022DiT}
William Peebles and Saining Xie.
\newblock Scalable diffusion models with transformers.
\newblock \emph{arXiv preprint arXiv:2212.09748}, 2022.

\bibitem[Poole et~al.(2022)Poole, Jain, Barron, and Mildenhall]{poole2022dreamfusion}
Ben Poole, Ajay Jain, Jonathan~T Barron, and Ben Mildenhall.
\newblock Dreamfusion: Text-to-3d using 2d diffusion.
\newblock \emph{arXiv preprint arXiv:2209.14988}, 2022.

\bibitem[Qi et~al.(2024)Qi, Yang, Zhang, Xing, Wu, Wu, Lin, Liu, Wang, and Zhao]{qi2024tailor3d}
Zhangyang Qi, Yunhan Yang, Mengchen Zhang, Long Xing, Xiaoyang Wu, Tong Wu, Dahua Lin, Xihui Liu, Jiaqi Wang, and Hengshuang Zhao.
\newblock Tailor3d: Customized 3d assets editing and generation with dual-side images.
\newblock \emph{arXiv preprint arXiv:2407.06191}, 2024.

\bibitem[Qu et~al.(2024)Qu, Xu, Hancke, and Lau]{qu2024lush}
Zefan Qu, Ke~Xu, Gerhard~Petrus Hancke, and Rynson~WH Lau.
\newblock Lush-nerf: Lighting up and sharpening nerfs for low-light scenes.
\newblock \emph{arXiv preprint arXiv:2411.06757}, 2024.

\bibitem[Qu et~al.(2025)Qu, Wang, Wang, Xu, Hancke, and Lau]{qu2025stylesculptor}
Zefan Qu, Zhenwei Wang, Haoyuan Wang, Ke~Xu, Gerhard~Petrus Hancke, and Rynson~WH Lau.
\newblock Stylesculptor: Zero-shot style-controllable 3d asset generation with texture-geometry dual guidance.
\newblock In \emph{Proceedings of the SIGGRAPH Asia 2025 Conference Papers}, pp.\  1--12, 2025.

\bibitem[Radford et~al.(2021)Radford, Kim, Hallacy, Ramesh, Goh, Agarwal, Sastry, Askell, Mishkin, Clark, et~al.]{radford2021learning}
Alec Radford, Jong~Wook Kim, Chris Hallacy, Aditya Ramesh, Gabriel Goh, Sandhini Agarwal, Girish Sastry, Amanda Askell, Pamela Mishkin, Jack Clark, et~al.
\newblock Learning transferable visual models from natural language supervision.
\newblock In \emph{International Conference on Machine Learning}, pp.\  8748--8763. PmLR, 2021.

\bibitem[Ren et~al.(2024)Ren, Huang, Zeng, Museth, Fidler, and Williams]{ren2024xcube}
Xuanchi Ren, Jiahui Huang, Xiaohui Zeng, Ken Museth, Sanja Fidler, and Francis Williams.
\newblock Xcube: Large-scale 3d generative modeling using sparse voxel hierarchies.
\newblock In \emph{Proceedings of the IEEE/CVF Conference on Computer Vision and Pattern Recognition}, pp.\  4209--4219, 2024.

\bibitem[Rombach et~al.(2022)Rombach, Blattmann, Lorenz, Esser, and Ommer]{rombach2022high}
Robin Rombach, Andreas Blattmann, Dominik Lorenz, Patrick Esser, and Bj{\"o}rn Ommer.
\newblock High-resolution image synthesis with latent diffusion models.
\newblock In \emph{Proceedings of the IEEE/CVF Conference on Computer Vision and Pattern Recognition}, pp.\  10684--10695, 2022.

\bibitem[Sabour et~al.(2025)Sabour, Goli, Kopanas, Matthews, Lagun, Guibas, Jacobson, Fleet, and Tagliasacchi]{sabour2025spotlesssplats}
Sara Sabour, Lily Goli, George Kopanas, Mark Matthews, Dmitry Lagun, Leonidas Guibas, Alec Jacobson, David Fleet, and Andrea Tagliasacchi.
\newblock Spotlesssplats: Ignoring distractors in 3d gaussian splatting.
\newblock \emph{ACM Transactions on Graphics}, 44\penalty0 (2):\penalty0 1--11, 2025.

\bibitem[Seed(2025)]{seed2025seed3d}
ByteDance Seed.
\newblock Seed3d 1.0: From images to high-fidelity simulation-ready 3d assets.
\newblock 2025.

\bibitem[Shue et~al.(2023)Shue, Chan, Po, Ankner, Wu, and Wetzstein]{shue20233d}
J~Ryan Shue, Eric~Ryan Chan, Ryan Po, Zachary Ankner, Jiajun Wu, and Gordon Wetzstein.
\newblock 3d neural field generation using triplane diffusion.
\newblock In \emph{Proceedings of the IEEE/CVF Conference on Computer Vision and Pattern Recognition}, pp.\  20875--20886, 2023.

\bibitem[Sim{\'e}oni et~al.(2025)Sim{\'e}oni, Vo, Seitzer, Baldassarre, Oquab, Jose, Khalidov, Szafraniec, Yi, Ramamonjisoa, et~al.]{simeoni2025dinov3}
Oriane Sim{\'e}oni, Huy~V Vo, Maximilian Seitzer, Federico Baldassarre, Maxime Oquab, Cijo Jose, Vasil Khalidov, Marc Szafraniec, Seungeun Yi, Micha{\"e}l Ramamonjisoa, et~al.
\newblock Dinov3.
\newblock \emph{arXiv preprint arXiv:2508.10104}, 2025.

\bibitem[Tan et~al.(2021)Tan, Wong, Wang, Manivasagam, Ren, and Urtasun]{tan2021scenegen}
Shuhan Tan, Kelvin Wong, Shenlong Wang, Sivabalan Manivasagam, Mengye Ren, and Raquel Urtasun.
\newblock Scenegen: Learning to generate realistic traffic scenes.
\newblock In \emph{Proceedings of the IEEE/CVF Conference on Computer Vision and Pattern Recognition}, pp.\  892--901, 2021.

\bibitem[Team et~al.(2024)Team, Anil, Borgeaud, Alayrac, Yu, Soricut, Schalkwyk, Dai, Hauth, Millican, et~al.]{team2024gemini}
Gemini Team, Rohan Anil, Sebastian Borgeaud, Jean-Baptiste Alayrac, Jiahui Yu, Radu Soricut, Johan Schalkwyk, Andrew~M Dai, Anja Hauth, Katie Millican, et~al.
\newblock Gemini: A family of highly capable multimodal models.
\newblock \emph{arXiv preprint arXiv:2312.11805}, 2024.

\bibitem[Team(2025)]{hunyuan3d2025hunyuan3d}
Tencent~Hunyuan3D Team.
\newblock Hunyuan3d 2.1: From images to high-fidelity 3d assets with production-ready pbr material, 2025.

\bibitem[Wang et~al.(2024)Wang, Wang, Hancke, Liu, and Lau]{wang2024themestation}
Zhenwei Wang, Tengfei Wang, Gerhard Hancke, Ziwei Liu, and Rynson~WH Lau.
\newblock Themestation: generating theme-aware 3d assets from few exemplars.
\newblock In \emph{ACM SIGGRAPH 2024 Conference Papers}, pp.\  1--12, 2024.

\bibitem[Wen et~al.(2025)Wen, Huang, Wang, Chen, and Sheng]{wen2025ouroboros3d}
Hao Wen, Zehuan Huang, Yaohui Wang, Xinyuan Chen, and Lu~Sheng.
\newblock Ouroboros3d: Image-to-3d generation via 3d-aware recursive diffusion.
\newblock In \emph{Proceedings of the Computer Vision and Pattern Recognition Conference}, pp.\  21631--21641, 2025.

\bibitem[Wu et~al.(2024)Wu, Lin, Zhang, Zeng, Xu, Torr, Cao, and Yao]{wu2024direct3d}
Shuang Wu, Youtian Lin, Feihu Zhang, Yifei Zeng, Jingxi Xu, Philip Torr, Xun Cao, and Yao Yao.
\newblock Direct3d: Scalable image-to-3d generation via 3d latent diffusion transformer.
\newblock \emph{arXiv preprint arXiv:2405.14832}, 2024.

\bibitem[Wu et~al.(2025)Wu, Lin, Zhang, Zeng, Yang, Bao, Qian, Zhu, Cao, Torr, et~al.]{wu2025direct3d}
Shuang Wu, Youtian Lin, Feihu Zhang, Yifei Zeng, Yikang Yang, Yajie Bao, Jiachen Qian, Siyu Zhu, Xun Cao, Philip Torr, et~al.
\newblock Direct3d-s2: Gigascale 3d generation made easy with spatial sparse attention.
\newblock \emph{arXiv preprint arXiv:2505.17412}, 2025.

\bibitem[Xiang et~al.(2024)Xiang, Lv, Xu, Deng, Wang, Zhang, Chen, Tong, and Yang]{xiang2024structured}
Jianfeng Xiang, Zelong Lv, Sicheng Xu, Yu~Deng, Ruicheng Wang, Bowen Zhang, Dong Chen, Xin Tong, and Jiaolong Yang.
\newblock Structured 3d latents for scalable and versatile 3d generation.
\newblock \emph{arXiv preprint arXiv:2412.01506}, 2024.

\bibitem[Xiang et~al.(2025)Xiang, Chen, Xu, Wang, Lv, Deng, Zhu, Dong, Zhao, Yuan, et~al.]{xiang2025native}
Jianfeng Xiang, Xiaoxue Chen, Sicheng Xu, Ruicheng Wang, Zelong Lv, Yu~Deng, Hongyuan Zhu, Yue Dong, Hao Zhao, Nicholas~Jing Yuan, et~al.
\newblock Native and compact structured latents for 3d generation.
\newblock \emph{arXiv preprint arXiv:2512.14692}, 2025.

\bibitem[Xu et~al.(2024)Xu, Mei, and Patel]{xu2024wild}
Jiacong Xu, Yiqun Mei, and Vishal Patel.
\newblock Wild-gs: Real-time novel view synthesis from unconstrained photo collections.
\newblock \emph{Advances in Neural Information Processing Systems}, 37:\penalty0 103334--103355, 2024.

\bibitem[Yang et~al.(2025)Yang, Lan, Chen, and Pan]{yang2025textured}
Songlin Yang, Yushi Lan, Honghua Chen, and Xingang Pan.
\newblock Textured 3d regenerative morphing with 3d diffusion prior.
\newblock \emph{arXiv preprint arXiv:2502.14316}, 2025.

\bibitem[Yin et~al.(2025)Yin, Cao, and Han]{yin2025wukong}
Minghao Yin, Yukang Cao, and Kai Han.
\newblock Wukong's 72 transformations: High-fidelity textured 3d morphing via flow models.
\newblock \emph{arXiv preprint arXiv:2511.22425}, 2025.

\bibitem[Zhang et~al.(2023)Zhang, Tang, Niessner, and Wonka]{zhang20233dshape2vecset}
Biao Zhang, Jiapeng Tang, Matthias Niessner, and Peter Wonka.
\newblock 3dshape2vecset: A 3d shape representation for neural fields and generative diffusion models.
\newblock \emph{ACM Transactions On Graphics (TOG)}, 42\penalty0 (4):\penalty0 1--16, 2023.

\bibitem[Zhang et~al.(2024)Zhang, Wang, Zhang, Qiu, Pang, Jiang, Yang, Xu, and Yu]{zhang2024clay}
Longwen Zhang, Ziyu Wang, Qixuan Zhang, Qiwei Qiu, Anqi Pang, Haoran Jiang, Wei Yang, Lan Xu, and Jingyi Yu.
\newblock Clay: A controllable large-scale generative model for creating high-quality 3d assets.
\newblock \emph{ACM Transactions on Graphics (TOG)}, 43\penalty0 (4):\penalty0 1--20, 2024.

\bibitem[Zhang et~al.(2018)Zhang, Isola, Efros, Shechtman, and Wang]{zhang2018perceptual}
Richard Zhang, Phillip Isola, Alexei~A Efros, Eli Shechtman, and Oliver Wang.
\newblock The unreasonable effectiveness of deep features as a perceptual metric.
\newblock In \emph{CVPR}, 2018.

\bibitem[Zhao et~al.(2025)Zhao, Lai, Lin, Zhao, Liu, Yang, Feng, Yang, Zhang, Yang, et~al.]{zhao2025hunyuan3d}
Zibo Zhao, Zeqiang Lai, Qingxiang Lin, Yunfei Zhao, Haolin Liu, Shuhui Yang, Yifei Feng, Mingxin Yang, Sheng Zhang, Xianghui Yang, et~al.
\newblock Hunyuan3d 2.0: Scaling diffusion models for high resolution textured 3d assets generation.
\newblock \emph{arXiv preprint arXiv:2501.12202}, 2025.

\bibitem[Zheng et~al.(2025)Zheng, Tan, Zhang, Wang, Guibas, Wetzstein, and Yifan]{zheng2025splatpainter}
Yang Zheng, Hao Tan, Kai Zhang, Peng Wang, Leonidas Guibas, Gordon Wetzstein, and Wang Yifan.
\newblock Splatpainter: Interactive authoring of 3d gaussians from 2d edits via test-time training.
\newblock \emph{arXiv preprint arXiv:2512.05354}, 2025.

\end{thebibliography}
\bibliographystyle{iclr2026_conference}

\newpage
\appendix
\section{Appendix}

\subsection{Motivation}
\begin{figure}[h]
  \centering
  \includegraphics[width=0.8\textwidth]{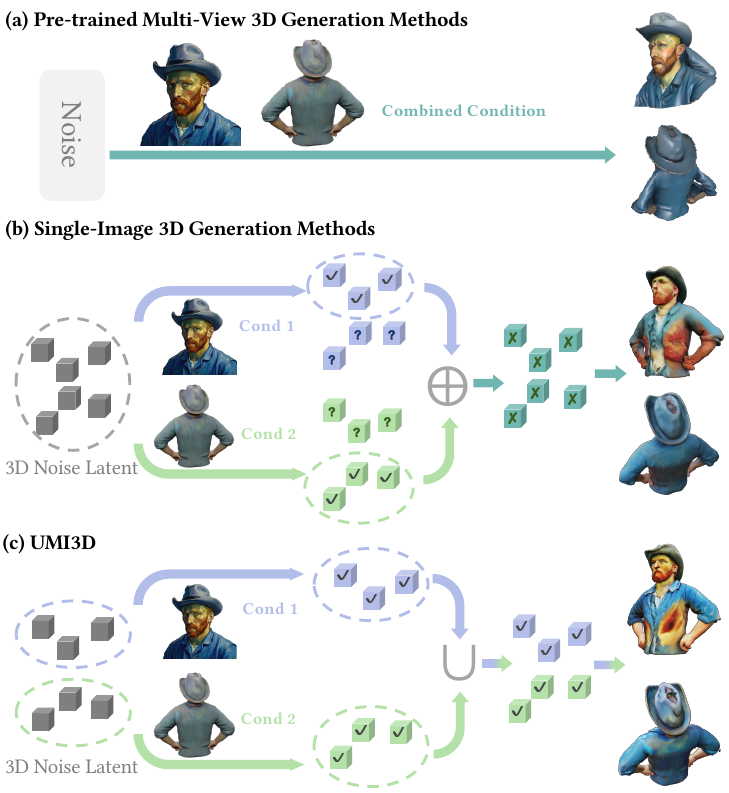}
  \caption{Limitations of existing multi-image 3D generation approaches under inconsistent inputs.}
  \label{fig:intro}
\end{figure}

The limitations of existing multi-image 3D generation approaches under inconsistent inputs are shown in the Fig.~\ref{fig:intro}. (a) Pre-trained multi-view 3D generation backbones assume relatively consistent inputs and fail under style or viewpoint mismatch. (b) Naive extensions of single-image 3D pipelines switch conditioning images across denoising steps, resulting in conditioning oscillation and detail loss. \ryn{(c) UMI3D resolves the above} issues via VRS-guided voxel routing.

\subsection{Preliminaries}\label{preliminary}
\subsubsection{Voxel-based 3D Generation Pipeline}
Voxel-based 3D generation pipelines~\citep{xiang2024structured,xiang2025native,wu2025direct3d,chen2025sam} serve as backbones for synthesizing 3D assets from user-provided images, and can produce multiple forms of 3D representations (\eg, NeRF, 3DGS, and meshes). Such pipelines typically consist of two or three stages, each responsible for generating a coarse 3D structure, detailed geometry, or texture, respectively. All stages are commonly implemented using rectified flow models~\citep{liu2022flow}. Since our method operates in a zero-shot manner, we focus only on the inference process.

In the first stage, the DINO model~\citep{oquab2023dinov2, simeoni2025dinov3} is used to extract image features \( c \) from the input images, which serve as conditions for the subsequent generation stages. A 3D noise feature grid \( S_t \) is then initialized and fed into the Flow Transformer \( V_{FT} \) for noise prediction:
\begin{equation}
S_{t-\Delta t} = S_t - V_{FT}(c, S_t, t) \, \Delta t \label{step1},
\end{equation}
where \( t \) and $\Delta t$ represent the time variable and time step length. After obtaining $ S_0 $, the decoder processes it to generate the voxels of the sparse structure of the asset, denoted as $ \{p_i\}_{i=1}^L $, where $p_i \in \mathbb{R}^3$ is the 3D coordinate of the $i$-th active voxel and $L$ is the total number of 
active voxels.
For each voxel, a noise latent $ \{Z_i\}_{i=1}^L $ is initialized and input into the Sparse Flow Transformer ($ V_{SFT} $) for the following stages of denoising:
\begin{equation}
Z_{t-\Delta t} = Z_t - V_{SFT}(c, p, Z, t) \, \Delta t \label{step2}.
\end{equation}

By processing this latent 3D representation through different decoders, the network can \ryn{then} generate various types of 3D asset \ryn{models}. Both $V_{\text{FT}}$ and $V_{\text{SFT}}$ adopt a DiT-based architecture~\citep{Peebles2022DiT} composed of multiple stacked transformer layers. 

\paragraph{Cross-attention conditioning.}
During 3D generation, the 2D information $c$ is injected into the generative network at \ryn{each layer via cross-attention (cross-attn)}. Given voxel tokens $\mathbf{Z} \in \mathbb{R}^{N_v \times d}$ and image features $c \in \mathbb{R}^{N_c \times d_c}$, the queries, keys, and values are computed as:
\begin{equation}\label{qkv}
\mathbf{Q} = \mathbf{Z}\mathbf{W}_Q,\quad
\mathbf{K} = c\,\mathbf{W}_K,\quad
\mathbf{V} = c\,\mathbf{W}_V,
\end{equation}
where $\mathbf{W}_Q$, $\mathbf{W}_K$, and $\mathbf{W}_V$ are learnable projection matrices. The cross-attn from voxel tokens to condition $c$ is then formulated as:
\begin{equation} \label{ca}
\operatorname{Cross-Attn}(\mathbf{Q}, \mathbf{K}, \mathbf{V})
= \operatorname{softmax}\!\left(
    \frac{\mathbf{Q}\mathbf{K}^\top}{\sqrt{d_q}}
  \right)\mathbf{V},
\end{equation} 
Through the cross-attn operation, the 3D latent features are enriched with 2D image information, enabling conditional 3D generation. 
To extend such 3D generative models to the multi-image setting, the latent 3D representation should be able to incorporate evidence from all conditioning images within each denoising step, rather than relying on a single fixed reference image throughout \ryn{the generation}. 

\subsection{Experiment Setup}
\subsubsection{Implementation Details}
UMI3D is built upon the TRELLIS.2 backbone~\citep{xiang2025native}. For all experiments, we use the publicly available \textit{TRELLIS.2-4B} model and its pre-trained weights provided in the official codebase. The output 3D asset is generated at $1024^3$ voxel resolution. \qzf{An anchor image is randomly selected as the sparse structure reference in our benchmark.} For the sparse shape generation stage, we use only the anchor image as conditioning during the first 75\% of denoising steps, and replace every \ryn{cross-attn} layer with SFC-Attn in the remaining 25\%. This \ryn{arrangement} preserves a stable coarse shape while allowing SFC-Attn to refine local geometry. For the subsequent structure and texture generation stages, we replace all the \ryn{cross-attn} layers, while keeping the rest of the network architecture unchanged. 
All model weights are frozen during inference, and no fine-tuning or parameter updates are required. 

The \ryn{scales of the classifier-free guidance (CFG)} are set to 7.5, 7.5, and 4.5 for the three stages, with 12 denoising steps per stage. For the 3D-KNN refine, the number of nearest neighbors $K_{\text{nn}}$ for each voxel is set to 10. All RGB views and normal maps reported in this paper are rendered from the generated 3D mesh under identical camera and lighting settings. All experiments are performed on a single NVIDIA RTX4090 GPU.

\subsubsection{Benchmark}
Since directly collecting multiple unconstrained images that depict exactly the same real-world object is impractical, we construct inconsistent multi-view inputs from consistent 3D assets in Sketchfab\footnote{https://sketchfab.com.} (CC BY 4.0). The selected assets cover diverse categories as well as a wide range of geometric and texture complexity.
Specifically, we select 40 diverse 3D models covering characters, animals, and everyday objects, and render each into 2-4 views with minimal overlap. 
We then use Gemini Nano-Banana-Pro~\citep{team2024gemini} to edit the renderings along one or more of six axes, including motion, style, occlusion, illumination, color, and local details, to produce semantically related but visually inconsistent views.
This process yields 40 unconstrained multi-image test cases, including 12 cases with 2 input images, 20 cases with 3 input images, and 8 cases with 4 input images. The edited images are manually checked to ensure semantic consistency with the original object while introducing realistic cross-view inconsistency. All test cases are fixed before evaluation to ensure a fair comparison across methods.

\begin{figure}[h]

  \centering
  \includegraphics[width=0.75\textwidth]{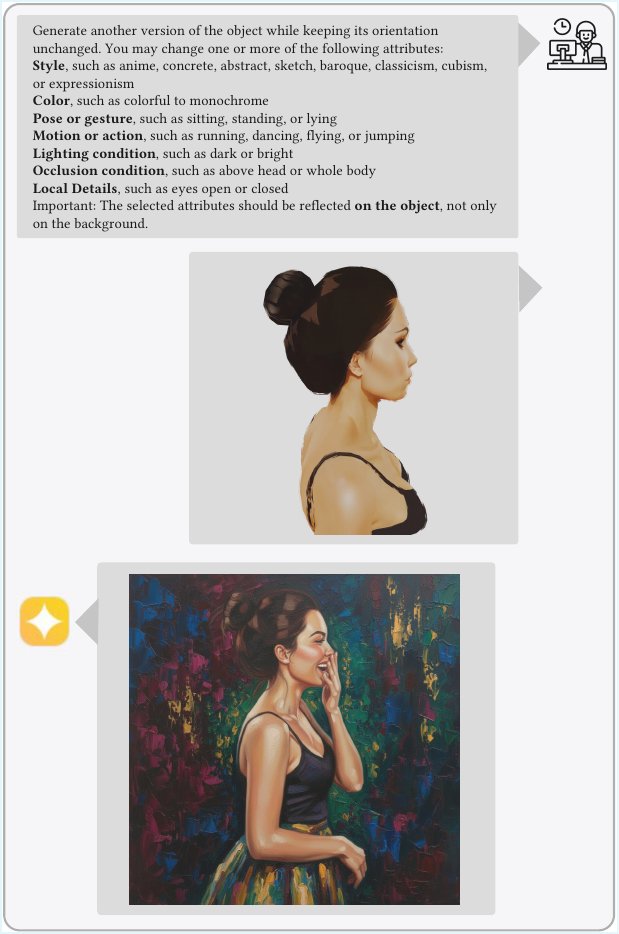}

  \caption{Example of the prompt to the image generation model.
  }
  \label{fig:prompt}

\end{figure}

The prompt to the Nano-Banana-Pro is illustrated in the Fig.~\ref{fig:prompt}. We manually filter out generated images that deviate significantly from the semantic information of the original objects (e.g. from human to cat). The dataset includes a wide range of inconsistencies, such as variations in lighting, motion, occlusion, color, style, and local details, ensuring the dataset’s diversity and the benchmark’s robustness.

\subsection{Metrics}
We adopt four metrics for quantitative evaluation:
(1) \textbf{CLIP}~\citep{radford2021learning}, which computes the cosine similarity between input images and rendered views using a pre-trained CLIP model;
(2) \textbf{DINO}~\citep{oquab2023dinov2}, which measures input-output cosine similarity using pre-trained DINOv2 features;
(3) \textbf{LPIPS}~\citep{zhang2018perceptual}, which evaluates perceptual distance between images;
and (4) \textbf{User Preference}, in which 29 participants complete a questionnaire comparing different methods over 20 randomly selected cases.

For CLIP, DINO, and LPIPS, we manually find the rendered views from the generated 3D asset that best match the viewpoints of the input images, and compute the metrics between these rendered views and the corresponding inputs. For fair comparison, all methods are evaluated under the same viewpoint selection and rendering settings. Since these image-based metrics mainly assess consistency at observed viewpoints rather than the quality of the complete 3D asset, we additionally conduct a user study using images rendered from more viewpoints to better evaluate overall user preference.

\subsubsection{Ablation Study Settings}
To evaluate the effectiveness of each component, we conduct an ablation study as shown in Fig.~\ref{fig:ablation}. The variants of our approach are \ryn{selected} as follows: 
(a) The Baseline Model (TRELLIS.2): Original \ryn{cross-attn} in both stages.
(b) w/o Geometry Routing: \ryn{Replacing} SFC-Attn in geometry generation with original cross-attn.
(c) w/o Texture Routing: \ryn{Replacing} SFC-Attn in texture generation with original cross-attn.
(d) w/o 3D-KNN: \ryn{Removing} the 3D-KNN refinement in SFC-Attn. 
(e) \ryn{Our} Full UMI3D Model.

\subsection{Extra Experiments}

\subsubsection{Generalization Ability}

\begin{figure}[h]
\centering
  \includegraphics[width=\textwidth]{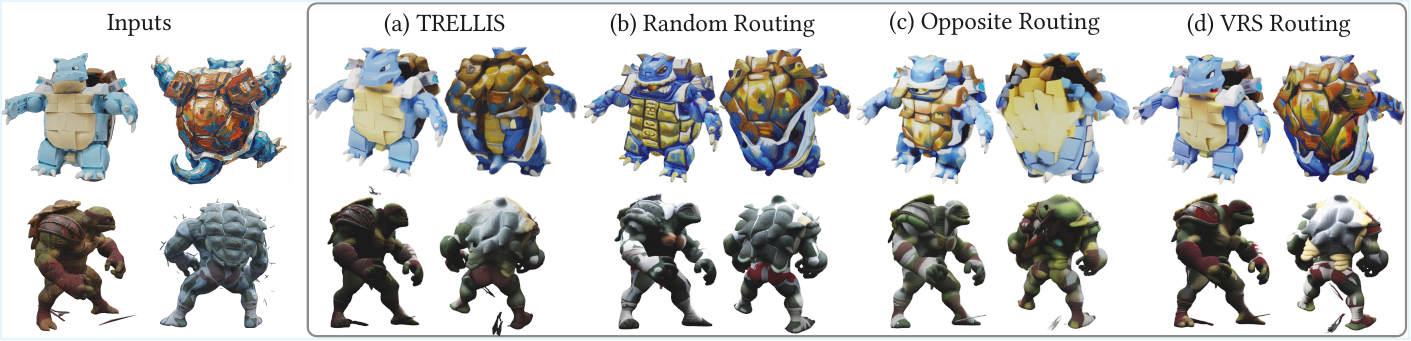}
  \caption{The generalization experiment on TRELLIS backbone.}
  \label{fig:general}
\end{figure}

To further demonstrate the generalizability of our method, we apply it to TRELLIS~\citep{xiang2024structured}. \ryn{Specifically}, we substitute the original cross-attn computation in the stage-2 voxel-based transformer layers with our proposed SFC-Attn module, while leaving the remaining architecture unchanged. The qualitative results are shown in Fig.~\ref{fig:general}.

As illustrated in \ryn{Fig.~\ref{fig:general}(a,d)}, our SFC-Attn module can be seamlessly integrated into the TRELLIS backbone and leads to improved 3D generation quality, especially for regions requiring stronger consistency across input views, such as the back of the blue cartoon character. In addition, \ryn{Fig.~\ref{fig:general}(b--d)} confirms that the insights presented in Sec.~\ref{sec:vrs} also extend to TRELLIS, with experimental observations consistent with those reported in Sec.~\ref{exp_insight}. Overall, these results indicate that UMI3D is compatible with other pre-trained 3D generation backbones and has promising generalization potential beyond the architecture studied in our main experiments.

\subsection{Applications}
\begin{figure*}[h]

  \includegraphics[width=\textwidth]{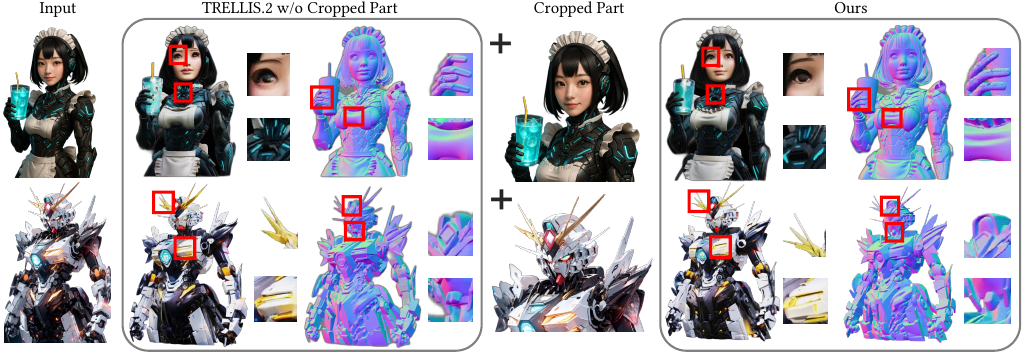}
  \vspace{-6mm}
  \caption{The single-view 3D generation refinement results of UMI3D.}
  \label{fig:single-view}
   \vspace{-2mm}
\end{figure*}
\paragraph{Single-View 3D Generation Local Enhancement.}
We use a mechanism \ryn{to refine} 3D generation results based on a single image by utilizing DINO features from a local image, which typically contain more detailed information than those from the entire image. As shown in Fig.~\ref{fig:single-view}, we manually crop the region \ryn{for refinement}, and input the corresponding local image along with the original image into UMI3D to obtain a 3D asset with refined local details. \ryn{We can see that the adjusted,} generated result is more accurate in terms of detail, while also preserving the quality of the uncropped regions.

\paragraph{Multi-View 3D Generation Enhancement.}
The VRS and SFC-Attn mechanisms enable UMI3D to generate models by referencing all input images, which is particularly important for 3D generation scenarios involving multi-view inputs. Fig.~\ref{fig:multi-view} shows some of the generated results. Compared to the TRELLIS.2 baseline, UMI3D can generate 3D assets with finer geometry and more accurate textures (such as the knight’s armor texture and the mouth shape of the SpongeBob character).

\paragraph{3D Morphing.}
With a slight modification to SFC-Attn, we find that UMI3D can perform high-quality 3D morphing, as illustrated in Fig.~\ref{fig:morphing}. Specifically, instead of using the $argmax$ operation in the SFC-Attn module to compare VRS scores, we directly inject conditions according to the VRS scores from the target-image branch, where the degree of injection is controlled by the desired morphing extent. The morphing process is achieved by progressively increasing the number of voxels conditioned on the target image.


For the sparse structure generation stage, \qzf{high-VRS voxels often correspond to strongly target-specific structural regions. Injecting them too early can abruptly impose the target's global shape.} So the condition injection to the voxels is performed in ascending order of VRS scores from the target image. This strategy enables gradual modifications starting from local and fine-grained regions, thereby preventing large-scale target semantics from being introduced too early in the morphing process, which could otherwise lead to the collapse of the overall 3D structure.
For the structure and texture generation steps, conditions are injected in descending order of VRS scores. In this way, the model can accurately enrich the already injected coarse spatial structure with detailed geometry and texture information. Using VRS scores in this manner, UMI3D can achieve smooth and coherent 3D morphing even between source and target images with drastically different semantics (e.g., the elephant to the toy plane shown in Fig.~\ref{fig:morphing}).

\begin{figure}[h]
\includegraphics[width=\textwidth]{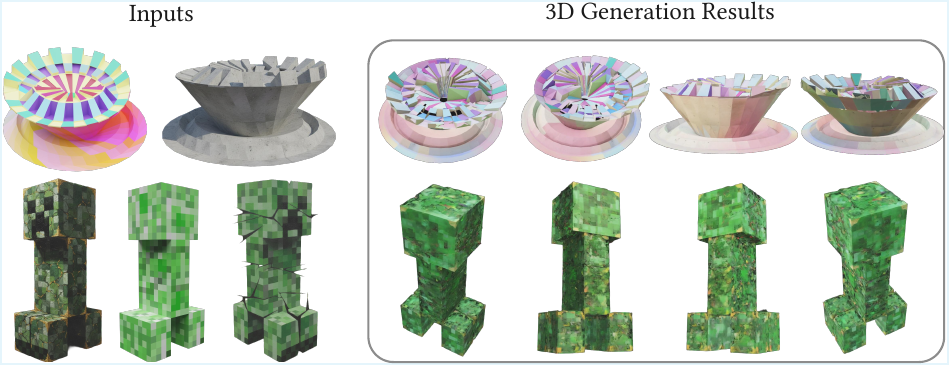}
  \caption{Failure Cases.
  }
  \label{fig:failure}
\end{figure}

\subsection{3D Assets Attribution.}

All 3D assets obtained from Sketchfab have been confirmed that have no \textit{NoAI} item. Each model used from Sketchfab in this paper is attributed as follows:
\begin{itemize}
\item[$\bullet$] “Portrait with Hat Van Gogh” by amicone.giuliana. 
\item[$\bullet$] "Blocky Jacket from Hotline Miami" by ILya Gamzayev.
\item[$\bullet$] "Toy Dragon" by Kai Christmann.
\item[$\bullet$] "Free - Male Armour - 3 - Game Ready" by Kaan Tezcan.
\item[$\bullet$] "Guilmon - Digimon World 4" by DrewsDigitalDesigns.
\item[$\bullet$] "Tower.Ruins.{*//X*}." by -X-ScornGames.
\item[$\bullet$] "Luffy" by MONKEY. D. LUFFY.
\item[$\bullet$] "tangerine peel man" by Sksskd.
\item[$\bullet$] "Swan" by Diana Liu.
\item[$\bullet$] "Colorful Frog Free" by moxstudios.
\item[$\bullet$] "owlbear-3d-model" by AidanYT55Twt.
\item[$\bullet$] "Wrecked" by Jacek Jaskólski.
\item[$\bullet$] "Mega Construx Pokémon (10)" by Emm.
\item[$\bullet$] "Teenage Mutant Ninja Turtles - Raphael
" by Hellbrush.
\item[$\bullet$] "Apple Watch Ultra 2" by polyman Studio.
\item[$\bullet$] "Knight - includes file for 3d printing" by Andy Woodhead.
\item[$\bullet$] "NASB2 - SpongeBob Cutscene" by SMF Features Developed From Cheryl Hill.
\item[$\bullet$] "a woman with a bunch of colorful bubbles in h" by klrxyz.
\item[$\bullet$] "Modern Art 008" by ParuthidotExE.
\item[$\bullet$] "Minecraft Creeper" by keithandmarchant.
\item[$\bullet$] "CatBus Mi Vecino Totoro" by Acalli Twiss.
\item[$\bullet$] "Creature" by tekovsky.
\item[$\bullet$] "A lying ox with a wreath of flowers" by Virtual Museums of Małopolska.
\item[$\bullet$] "REANIMAL - The Boy" by MG Rips.
\item[$\bullet$] "Mr. Mime in the Box - Practical Joke" by kuayarts.
\item[$\bullet$] "Lego Harry Potter: Professor Quirrell (2018)
3D Model" by Zorg\_Sinister.
\item[$\bullet$] "Scarlet Blade Heroine Rigged" by Ar3Designer.
\item[$\bullet$] "genshin impact ganyu Twilight Blossom" by jacenrick.
\item[$\bullet$] "Old Couch" by Oliver Triplett.

\end{itemize}

All 2D images used in this paper are selected from the images provided in the official TRELLIS~\citep{xiang2024structured} or Direct3D-S2~\citep{wu2025direct3d} code repositories. All of these images are AI generated and open source for the academic research purposes.

\begin{figure*}
  \includegraphics[width=\textwidth]{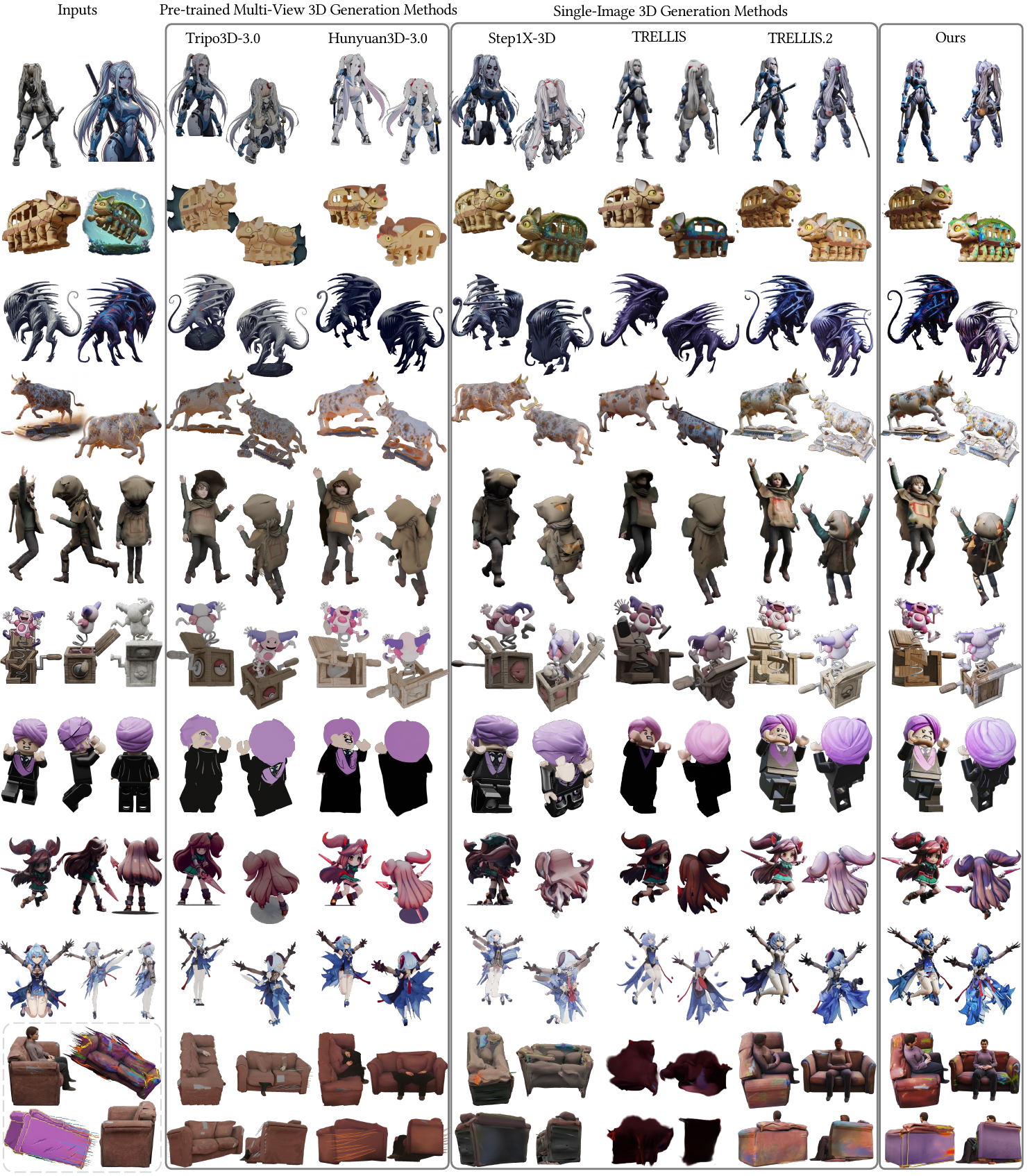}
  \caption{More qualitative comparison with other 3D generation methods.
  }
  \label{fig:comparison_all}
\end{figure*}

\end{document}